# Integrating Generative AI-enhanced Cognitive Systems in Higher Education: From Stakeholder Perceptions to a Conceptual Framework considering the EU AI Act

Da-Lun Chen*, Prasasthy Balasubramanian, Lauri Lovén,
Susanna Pirttikangas, Jaakko Sauvola, Panagiotis Kostakos

Faculty of Information Technology and Electrical Engineering,
University of Oulu, Oulu, Finland

*Corresponding author email: da-lun.chen@oulu.fi

## Abstract

Many staff and students in higher education have adopted generative artificial intelligence (GenAI) tools in their work and study. GenAI is expected to enhance cognitive systems by enabling personalized learning and streamlining educational services. However, stakeholders' perceptions of GenAI in higher education remain divided, shaped by cultural, disciplinary, and institutional contexts. In addition, the EU AI Act requires universities to ensure regulatory compliance when deploying cognitive systems. These developments highlight the need for institutions to engage stakeholders and tailor GenAI integration to their needs while addressing concerns. This study investigates how GenAI is perceived within the disciplines of Information Technology and Electrical Engineering (ITEE). Using a mixed-method approach, we surveyed 61 staff and 37 students at the Faculty of ITEE, University of Oulu. The results reveal both shared and discipline-specific themes, including strong interest in programming support from GenAI and concerns over response quality, privacy, and academic integrity. Drawing from these insights, the study identifies a set of high-level requirements and proposes a conceptual framework for responsible GenAI integration. Disciplinary-specific requirements reinforce the importance of stakeholder engagement when integrating GenAI into higher education. The high-level requirements and the framework provide practical guidance for universities aiming to harness GenAI while addressing stakeholder concerns and ensuring regulatory compliance.

## 1 Introduction

The adoption of ChatGPT and other generative artificial intelligence (GenAI) tools has been unprecedented. In just five days after its release, ChatGPT gained over one million users, marking a significant milestone in technology adoption (Mittal et al., 2024). Besides ChatGPT, other GenAI tools like Claude (text generation), Stable Diffusion (image generation), VoiceBox (audio generation), Sora (video creation), and GitHub Copilot (coding assistance) have entered the market. These tools, powered by deep learning models trained on immense datasets, share the unique trait of generating new content based on user prompts, distinguishing themselves from other AI branches.

This phenomenon has garnered global attention, prompting organizations and governments to take measures addressing the challenges and opportunities AI poses in general. For instance, (UNESCO, 2022) issued recommendations promoting AI literacy and ethics for different policy areas. The EU passed the AI Act aiming to regulate AI systems according to their risk categories (European Commission, 2025), which may influence how universities adopt GenAI. These initiatives underscore the transformative and disruptive impacts of AI on society.

In higher education, while many people have become aware of GenAI and started leveraging the technology in learning, teaching, and administration, the perceptions regarding the technology are mixed (Hashmi & Bal, 2024; Yusuf et al., 2024). On one end of the spectrum, enthusiasts embrace its transformative potential, believing its concerns can be addressed properly (Alharbi, 2024). On the other end, sceptics highlight unresolved issues such as the misuse of GenAI for academic misconduct and insufficient preparedness for institutional adoption. In between, there are people who perceive both the opportunities and challenges GenAI presents (Chan & Hu, 2023; Kasneci et al., 2023). Additionally, cultural factors can influence perceptions (Yusuf et al., 2024). These diverse perceptions reflect the complexity of integrating GenAI into higher education.

From a technical perspective, GenAI can significantly enhance cognitive systems that transform higher education, systems that provide smart applications leveraging cognitive computing and data for educational contexts, facilitating personalized adaptive learning and empowering decision making (Lytras et al., 2019). Educational chatbots are such applications under rapid transformation towards adopting GenAI. Many recent empirical studies focusing on language learning and short-term interventions have reported significant effects of ChatGPT (Deng et al., 2025). These initial results call for further research on crucial areas, including improving the accuracy (Labadze et al., 2023; Lo et al., 2024), ensuring AI literacy (Ansari et al., 2023; Labadze et al., 2023; Maheshwari, 2023), updating assessment methods (Deng et al., 2025), and integration that promotes higher-order thinking (Borge et al., 2024; Lo et al., 2024; Urban et al., 2024).

Given the ethical, pedagogical, and technological considerations surrounding GenAI-enhanced cognitive systems in higher education, integration must be approached holistically. This includes engaging stakeholders, aligning with institutional objectives, and ensuring regulatory compliance throughout the system's life cycle. This study addresses the challenges of integrating GenAI into higher education by:

- Examining faculty and student perceptions of GenAI to understand their views and concerns.

- Identifying high-level requirements from survey data to inform future design and implementation.

- Proposing a conceptual framework for integrating GenAI and other cognitive systems into higher education that meets stakeholders' needs, supports governance, and addresses regulatory requirements.

The subsequent sections are structured as follows: we begin by examining the current discourse surrounding GenAI in higher education, the evolving landscape of educational chatbots, and critical considerations for integrating GenAI-enhanced cognitive systems. Subsequently, we present empirical survey findings on faculty and student perceptions, identifying relevant high-level requirements for GenAI adoption. Finally, we propose a conceptual framework for integration, discuss its implications, and outline directions for future research.

# 2 Related Work

## 2.1 Mixed Perceptions of GenAI in Higher Education

GenAI refers to deep learning models that can produce new content in various formats beyond text. In their review, Bengesi et al. (2024) highlighted state-of-the-art GenAI architectures, including transformers, variational autoencoders (VAE), generative adversarial networks (GAN), and diffusion models. Transformers, particularly large language models (LLM) of generative pre-trained transformer (GPT), excel in natural language processing. In contrast, VAE models are often applied in simpler image processing tasks such as feature extraction or image reconstruction. GAN and diffusion models, on the other hand, are highly effective at generating high-quality images, videos, and audio.

In the context of education, researchers have anticipated that GenAI can engage students, enhance personalized and adaptive learning, increase efficiency, and foster creativity and innovation (Kasneci et al., 2023; Mittal et al., 2024). However, while we have witnessed the rapid adoption of GenAI on a global scale, its integration in education remains slower (Mittal et al., 2024). Perceptions of GenAI in higher education are particularly mixed among students and staff, as the technology is seen as presenting both opportunities and threats (Alharbi, 2024; Chan & Hu, 2023; Hashmi & Bal, 2024; Kasneci et al., 2023; Watermeyer et al., 2023; Yusuf et al., 2024). Additionally, cultural factors could further influence these perceptions (Yusuf et al., 2024), increasing the complexity of GenAI integration. These mixed perceptions are elaborated below.

### 2.1.1 Students' Perspective

Several recent studies suggest that many students in higher education have become aware of and have used GenAI through proprietary tools like ChatGPT (Bernabei et al., 2023; Chan & Hu, 2023; Jin et al., 2025; Neyem et al., 2024; Yusuf et al., 2024). They perceive benefits of GenAI such as streamlining information search and analysis (Chan & Hu, 2023), enhancing academic writing (Bernabei et al., 2023; Chan & Hu, 2023; Jin et al., 2025), creating multimedia content efficiently (Chan & Hu, 2023), providing assistance and guidance in software development projects (Neyem et al., 2024), and enabling 24/7 support and personalized learning (Chan & Hu, 2023).

On the other hand, students have also raised concerns regarding GenAI. In terms of the technology itself, concerns have been raised on its accuracy, transparency, biases, and risks of violating privacy, copyright and intellectual property (Bernabei et al., 2023; Chan & Hu, 2023; Jin et al., 2025). From a learning perspective, some students are concerned about diminishing critical thinking and creativity due to over reliance on the technology (Chan & Hu, 2023; Jin et al., 2025), and how it might affect negatively on the student-teacher relationship (Chan & Hu, 2023). Looking beyond their education, some students have also expressed worry regarding their career prospects (Bernabei et al., 2023; Chan & Hu, 2023) and how the technology might exacerbate social injustice and inequality (Chan & Hu, 2023). Last but not the least, students have articulated the need for comprehensive institutional policies and guidelines (Bernabei et al., 2023; Chan & Hu, 2023).

### 2.1.2 Staff's Perspective

Other recent studies also show that many staff members have become aware of GenAI and used it primarily for enhancing teaching and streamlining administrative tasks; for example, designing teaching plans and materials (Zaim et al., 2024), creating and supporting student assessments (Lee et al., 2024), generating multimedia contents (Cervantes et al., 2024), and supporting administrative and routine tasks (Cervantes et al., 2024; Lee et al., 2024; Watermeyer et al., 2023). From a positive perspective, staff members can see the transformative potential of GenAI, such as for enhancing their communication to their students (Garcia-Varela et al., 2025), streamlining student assessments and tailoring feedback to individual students (Tzirides et al., 2024), offering simulation-based learning (Cervantes et al., 2024), fostering inclusive learning environments (Ulla et al., 2024), facilitating personalized learning (Alharbi, 2024; Ulla et al., 2024), and supporting research (Cervantes et al., 2024).

However, despite the positive experiences and perceptions, the accuracy of GenAI has been one of the main concerns shared by staff from different higher education institutions (Alharbi, 2024; Cervantes et al., 2024; Lee et al., 2024). In addition, it has been noted that GPT models sometimes produce lengthy but general responses (Tzirides et al., 2024) or responses lacking subject context (Cervantes et al., 2024), highlighting the need to customize the tools. Another shared concern is the risk of compromising academic integrity due to abusing the technology. From a teaching perspective, some teachers have voiced their concern that GenAI may impede students' learning if they become overly dependent on the technology (Lee et al., 2024); furthermore, many have expressed concern about students' misuses of GenAI such as cheating and plagiarism (Alharbi, 2024; Cervantes et al., 2024; Lee et al., 2024; Ulla et al., 2024), leading to some teachers modifying their assessment methods (Lee et al., 2024). Similarly, critical views on the risks of GenAI degenerating research integrity, researcher's identity, and well-being have also been presented (Watermeyer et al., 2023). Other articulated concerns include, for example, privacy (Cervantes et al., 2024; Ulla et al., 2024), data protection (Ulla et al., 2024), and diminishing human interaction from important processes (Lee et al., 2024). These concerns and issues emphasize the importance of adequate AI policies, guidelines, staff training and support (Alharbi, 2024; Cervantes et al., 2024; Lee et al., 2024; Ulla et al., 2024; Zaim et al., 2024), as well as the need to carefully integrate GenAI, ensuring the use of technology aligning with the requirements and expected quality in higher education (Ulla et al., 2024; Zaim et al., 2024).

## 2.2 The Transformation of Educational Chatbots

Chatbots are software programs that interact with their users on specific subjects by imitating human-like conversation (Coniam, 2008; Weizenbaum, 1983). They either offer unimodal interaction, using text or voice, or support multimodal interaction (Jeon et al., 2023). In terms of how chatbots respond to users, they can be categorized as rule-based, intent-based, or generative (Kuhail et al., 2023). Rule-based chatbots follow predefined logic such as pattern matching and conditional dialog paths when interacting with users. Intent-based chatbots leverage natural language processing, sometimes combining rule-based logic, to identify the intent and entities in users' questions and present a predefined response accordingly. Generative chatbots, leveraging GenAI, dynamically tailor their responses to their users' questions.

The research and development of chatbots have come a long way. For example, ELIZA, often credited as the world's first chatbot (Alemdag, 2023; Coniam, 2008; Lin & Yu, 2023; Smutny & Schreiberova, 2020), was a rule-based unimodal chatbot developed by Joseph Weizenbaum at MIT and introduced in 1966 (Weizenbaum, 1983). It mimicked a psychiatrist in conversation by using pattern matching and scripted sentence modifications to create responses (Weizenbaum, 1983). Fast forward to 2011, the IBM Watson made headlines after defeating two former winners of a popular quiz show, Jeopardy!, using DeepQA, its advanced natural language processing technology, to analyse the questions and find the most plausible answer from its knowledge base (Lally & Fordor, 2011). Eleven years later, ChatGPT, a generative chatbot by OpenAI made its debut in November 2022 (OpenAI, 2022) and started a global sensation on GenAI's capabilities beyond natural language interaction.

Prior to ChatGPT, most reported educational chatbots were rule- or intent-based (Kuhail et al., 2023), designed for students and for specific purposes (Alemdag, 2023; Jeon et al., 2023). Compared to other application domains, they were frequently utilized in language learning (Alemdag, 2023; Jeon et al., 2023; Koc & Savas, 2025; Kuhail et al., 2023; Lin & Yu, 2023), computer science (Kuhail et al., 2023), general educational services, and healthcare education (Lin & Yu, 2023). These chatbots often served as teaching assistants, aiming to facilitate personalized learning (Alemdag, 2023; Jeon et al., 2023; Kuhail et al., 2023), provide scaffolding (Jeon et al., 2023; Koc & Savas, 2025; Kuhail et al., 2023), or support collaborative learning (Kuhail et al., 2023). Several reviews suggest that educational chatbots could engage students, enhance learning, and increase satisfaction (Alemdag, 2023; Jeon et al., 2023; Koc & Savas, 2025; Kuhail et al., 2023; Wu & Yu, 2024). However, it has also been observed that greater effects were often associated with shorter interventions (Alemdag, 2023; Wu & Yu, 2024), suggesting that these educational chatbots might be more suitable for short-duration use cases—ranging from a few sessions to several weeks. While rule- and intent-based chatbots were dominant at the time, efforts were also made to train and use generative chatbots, but these faced challenges due to insufficient datasets or inadequate training (Kuhail et al., 2023; Labadze et al., 2023).

In contrast to their predecessors, ChatGPT and other recent generative chatbots have demonstrated superior capabilities not only in natural language interaction but also in creating different kinds of new content based on the user's request. These advanced chatbot applications have reached various stakeholder groups in higher education through either a free version or a subscription. For example, some researchers explored ChatGPT for planning, refining research design, and polishing writing (Ansari et al., 2023). Teachers used it to streamline teaching and other work (Labadze et al., 2023). Many students used it to help them study, complete assignments, receive personalized feedback (Deng et al., 2025; Labadze et al., 2023; Lo et al., 2024; Ma et al., 2024), or practice English in their own time (Liu et al., 2024). These various uses highlight the versatile capabilities of these chatbots.

Meanwhile, research has started examining the educational effects of this new generation of chatbots in different learning contexts, including English learning, academic writing, problem solving, and collaborative learning. In language learning, early evidence suggests significant benefits of generative chatbots for engaging English-as-a-Second-Language students (Pan et al., 2025; Wang & Xue, 2024) and improving their oral proficiency (Guan et al., 2024). In the context of developing writing skills, generative chatbots can help students develop logical thinking, improve their understanding of argumentative writing, and identify areas for improvement (Zhang et al., 2025). They can also assist students in evaluating each other's text and improving their feedback and writing (Guo et al., 2024). Intriguingly, mixed results have been reported regarding problem solving and collaborative learning. Stadler et al. (2024) found that ChatGPT, while it reduced students' cognitive load in information search, did not help students make more informed decisions than those using a search engine. On the other hand, Urban et al. (2024) reported that ChatGPT empowered students to come up with more original, elaborated, and better solutions for complex problems while increasing their self-efficacy. Similarly, in the context of collaborative learning, Hu et al. (2025) did not find that generative chatbots improved the outcome of a collaborative writing activity, whereas An et al. (2025) reported enhanced collaboration in document composition. These mixed findings suggest that how generative chatbots are integrated could influence the outcomes, highlighting the need for more empirical studies. To conclude, Deng et al. (2025) found in their recent review on the effects of ChatGPT that most empirical studies reported significant improvements in academic performance, motivation, higher-order thinking, and reducing cognitive load. They further noted that ChatGPT had stronger effects in language learning and that short interventions (1-4 weeks) were more effective than longer ones.

As presented, ChatGPT and recent generative chatbots are changing the way chatbots are used in higher education—from chatbot-led to user-led interaction and from serving specific to general purposes, presenting unprecedented opportunities while also raising concerns, as elaborated in the mixed perceptions of GenAI in higher education. The applications and effects of generative chatbots call for further empirical research in several crucial areas, including improving their accuracy (Labadze et al., 2023; Lo et al., 2024), ensuring staff and students have access to up-to-date institutional policies, guidelines, and support (Ansari et al., 2023; Labadze et al., 2023; Maheshwari, 2023), updating assessment methods (Deng et al., 2025), and exploring effective ways to integrate generative chatbots into higher education to promote higher-order thinking (Borge et al., 2024; Lo et al., 2024; Urban et al., 2024).

## 2.3 Key Factors Influencing the Integration of GenAI into Cognitive Systems in Higher Education

Cognitive systems in higher education refer to the integration and effective use of big data, AI, and modern technologies that facilitate applications including personalized adaptive learning and data-driven decision making (Bahassi et al., 2024; Lytras et al., 2019; Rahman, 2022). For example, Zhuhadar and Ciampa (2019) demonstrated how cognitive computing can help learners derive insights from massive data. Aljohani et al. (2019) presented a cognitive system that allowed instructors to configure learning analytics for their online courses and provided students with personalized analytics feedback to enhance their engagement and performance. Capuano and Toti (2019) presented a smart learning system for law that leverages natural language processing, knowledge discovery, and adaptive learning paths, to help learners study legal concepts effectively. Acknowledging the transformative opportunities of GenAI and the need for safeguarding privacy, ensuring security and regulatory compliance, Ng et al. (2025) reported the implementation of a secured environment enabling medical researchers to investigate the applications of GenAI in fall detection and psychological assessment.

These systems are often composed of three common layers: the data layer, the cognitive computing layer, and the application layer (Bahassi et al., 2024). The data layer is responsible for collecting, storing, and preprocessing data relevant to the system, facilitating the next layer. The cognitive computing layer manages the lifecycle of various AI technologies, enabling advanced analytics, prediction and inference. Finally, the application layer develops and manages applications that leverage cognitive computing to deliver user-facing services. While these layers consist of the main building blocks of most cognitive systems, specific systems may include additional layers or components depending on their design and purpose. Furthermore, from a compositional perspective, individual systems can function as independent agents. These agents can be orchestrated into complex systems to deliver sophisticated services, highlighting not only the need for interoperability (Airaj, 2024) but also the challenge of addressing the requirements of different stakeholders.

While the accelerated advancement of GenAI offers opportunities to enhance cognitive systems and transform higher education, merely adopting the technology is not sufficient to realize anticipated benefits. As presented in the mixed perceptions of GenAI in higher education, students and faculty have expressed their concerns regarding ethics (e.g., privacy, bias, and academic integrity) and pedagogical aspects (such as impeding higher-order thinking). GenAI and other cognitive computing initiatives that do not adequately involve their stakeholders and address their concerns are likely to encounter additional resistance (Al-Omari et al., 2025). Additionally, evolving technology development and regulatory requirements pose further challenges for integrating cognitive systems into higher education (Balan, 2024). These aspects call attention to AI literacy, governance mechanisms, and collaboration among stakeholders.

Adequate AI literacy possessed by staff and students is fundamental to successful integration (Chan, 2023). Using Bloom's taxonomy as an analogy, AI literacy begins with obtaining general knowledge of AI and the skills of using AI applications (Ng et al., 2021). Building on the basic knowledge and skills, individuals can develop the competence to critically assess applications, considering not only their objectives but also ethical aspects. While technical expertise is further required for developing cognitive systems in higher education, staff and students should be equipped with the competence to evaluate these systems from the perspectives of their roles. Having this level of AI literacy, they can not only use GenAI and other cognitive systems responsibly but also contribute to the governance of such integration. Adequate AI literacy can be achieved through institutional training and support (Chan, 2023; Yang et al., 2024).

Effective governance can ensure that the integration of cognitive systems into higher education aligns with intended objectives, addresses ethical concerns, and complies with relevant policies and regulations. It requires establishing institutional policies that can guide responsible and ethical actions, while being adaptable for future changes (Chan, 2023; Hughes et al., 2025) and providing instructors with sufficient space to specify whether and how AI is used in their classes (Dabis & Csáki, 2024). Committees should be set up strategically to offer advisory services to AI initiatives and perform regular evaluations on the performance and impacts of integrated systems (Balan, 2024). Effective governance also requires universities to stay informed about regulatory developments and ensure that their cognitive systems comply with regulations. Besides national regulations, it is also imperative for those universities with international reach to consider the policies with extraterritorial implications such as GDPR and the EU AI Act. Like the GDPR, the EU AI Act has extraterritorial implications, meaning that AI developers and providers outside the EU must comply if their systems impact individuals or business within the Union (European Commission, 2025).

As defined by the EU AI Act, cognitive systems are classified into four distinct risk levels: Unacceptable Risk, High Risk, Limited Risk, and Minimal Risk. These levels are depicted in Fig. 1. Systems that clearly endanger people's safety and rights fall into the Unacceptable Risk category and are prohibited by the Act (e.g., systems performing "social scoring" on campuses). The High Risk category includes systems which could have significant consequences on human lives or violate human rights if they fail to function properly (e.g., cognitive systems for streamlining degree program admission or course enrolment). These systems are required to meet additional compliance requirements before being introduced to the market. Major modifications to these systems in the future will require reassessment to ensure continued compliance. In contrast, systems that leverage GenAI and do not fall under the previous categories are considered Limited Risk. However, their providers must make it clear that content is AI-generated (e.g., a GenAI chatbot must inform users that they are interacting with AI). Lastly, systems that do not fall into any of the aforementioned categories (e.g., AI-enabled video games or spam filters) are not subject to specific rules under the AI Act.

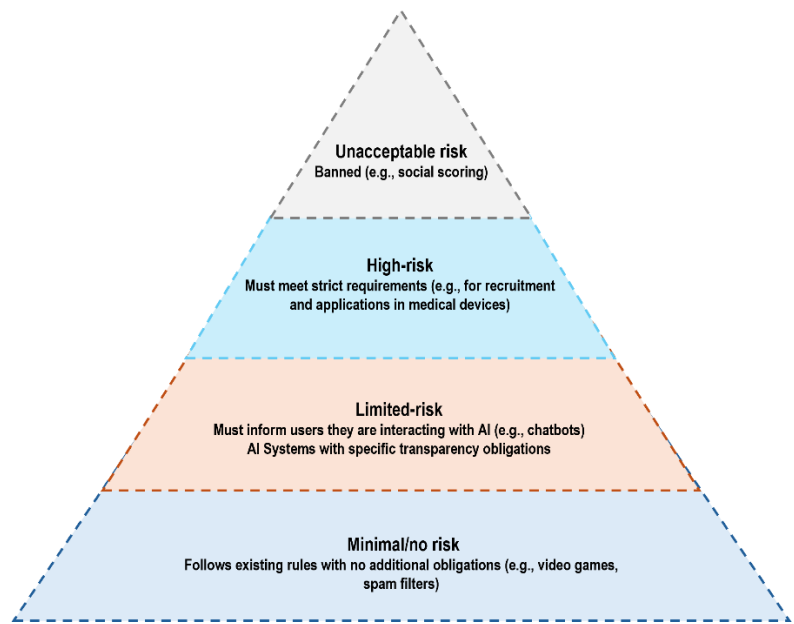

*Fig. 1 AI-Powered Educational Platform (Balasubramanian et al., 2025)*

Adequate AI literacy and effective governance, alongside technical development, are thus essential for successfully integrating cognitive systems into higher education, underscoring the need to identify and engage stakeholders in various collaborations. To promote adequate AI literacy across the institution, staff and students should be engaged in institutional training and provided with support (Chan, 2023). Further incorporating trustworthy AI into the curriculum necessitates interdisciplinary collaboration (Aler Tubella et al., 2023). In terms of governance, joint effort involving staff and students has been advocated for creating policies (Chan, 2023; Hughes et al., 2025), establishing committees (Burk-Rafel & Triola, 2025), and evaluating the impacts of cognitive systems (Dabis & Csáki, 2024). Aligning with the intended objectives and complying with policies and regulations necessitate proactive engagement of relevant stakeholders in technical development (Mangal & Pardos, 2024). Beyond staff and students within the institution, external experts, regulatory bodies, and technology firms can also be important stakeholders in the integration process, depending on the scope of the initiative (Aler Tubella et al., 2023; Al-Omari et al., 2025; Chan, 2023; Pachava et al., 2025). These governance- and stakeholder-related considerations form a foundation for successful integration.

# 3 Design of GenAI Staff and Student Surveys

To explore and compare the perspectives of staff and students on GenAI and to identify high-level requirements for the Faculty of Information Technology and Electrical Engineering at the University of Oulu, we developed a staff survey and a student survey with similar or identical sections. Both surveys collected demographic data and responses on the following areas: perceptions of educational chatbots in higher education, use of chatbots and other GenAI in work or study, perceived impacts, concerns, and ideas for GenAI applications in higher education. As our study did not focus on socio-economic aspects, we limited demographic questions to gender, age range, and job titles in the staff survey, while the student survey included gender, age range, study status, and access to a computer and the internet. Most survey items were ordinal or categorical, with some open-ended questions to collect additional insights. During the survey design phase, AI tools were used for reviewing survey items (MS Copilot) and translating the survey content into Finnish (MS Office Translator). Subsequently, two pilots were carried out with a small group of staff members, including research assistants who were also students, to refine the surveys.

After piloting, the surveys were administered using an online survey tool subscribed to by the university. Data were collected in May 2024 through convenience sampling. Separate email invitations with respective survey links were sent to faculty staff and students via their mailing lists. To introduce GenAI to participants with little or no prior experience, a GenAI chatbot was set up and linked in the survey introduction. Additionally, the staff survey was promoted through the faculty newsletter and Teams space, while the student survey was promoted through guild visits and printed fliers. Participation in both surveys was anonymous and voluntary, adhering to established ethical practices (informed consent principles) and guidelines issued by the University of Oulu. The English versions of both the staff and student surveys are provided as supplementary information.

Originally, we collected 61 responses from the staff survey and 38 from the student survey. Most staff respondents held teaching or research roles, such as professors, lecturers, doctoral researchers, and research assistants. Among student respondents, the majority were bachelor's or master's students, with one doctoral student included. It is worth noting that students may also hold staff roles such as research assistants or doctoral researchers. To enable statistical comparison between staff and student responses using methods such as the Mann–Whitney U test, a sensitivity analysis was conducted to examine whether excluding research assistants from the staff dataset and the single doctoral student from the student dataset would affect the test results. The analysis showed that excluding research assistants did not influence the results, while excluding the doctoral student reduced one observed statistical difference. As there was no additional information to confirm that the doctoral student also held a staff role, and to avoid violating the assumptions of the Mann–Whitney U test or risking a false positive, we decided to exclude this response from the student dataset. Consequently, the final dataset used for analysis included 61 staff and 37 student responses. In addition, open-ended responses were analysed using an iterative thematic analysis approach.

# 4 Survey Results

## 4.1 Demographics

Most of the staff respondents were male (69%, n=61), with the age range 40-49 being reported most frequently (33%). While the majority had teaching and research roles, doctoral researcher was the most frequently reported job title (33%) (see Table 1). At work, staff were provided with work machines and access to high-speed Internet.

*Table 1 Demographics of Staff Respondents (n=61)*

| Variable | count | % |
| --- | ---: | ---: |
| **Gender** | | |
| Female | 13 | 21 |
| Male | 42 | 69 |
| Prefer not to say | 3 | 5 |
| No response | 3 | 5 |
| **Age range** | | |
| 20-29 | 6 | 10 |
| 30-39 | 16 | 26 |
| 40-49 | 20 | 33 |
| 50-59 | 12 | 20 |
| 60-69 | 4 | 7 |
| Prefer not to say | 3 | 5 |
| **Job title(s)** | | |
| Professor | 7 | 12 |
| Associate Professor | 1 | 2 |
| Assistant Professor | 2 | 3 |
| Postdoctoral Researcher | 5 | 8 |
| Lecturer | 12 | 20 |
| Project manager | 7 | 12 |
| Administrative staff | 5 | 8 |
| Doctoral researcher | 20 | 33 |
| Research assistant | 5 | 8 |
| Other | 1 | 2 |

Among the student respondents, 70% (n=37) were male and 62% were aged between 20-29. Just over half were studying for a bachelor's degree (51%), while the rest were pursuing a master's degree (49%). Nearly half had been studying for less than a year at the time of the survey (49%). Lastly, all student respondents had access to a personal computer or laptop, and almost all had high-speed internet access (97%) (See Table 2).

*Table 2 Demographics of Student Respondents (n=37)*

| Variable | count | % |
|---|---|---|
| **Gender** | | |
| Female | 9 | 24 |
| Male | 26 | 70 |
| Other | 1 | 3 |
| Prefer not to say | 1 | 3 |
| **Age range** | | |
| <20 | 2 | 5 |
| 20-29 | 23 | 62 |
| 30-39 | 9 | 24 |
| 40-49 | 3 | 8 |
| **Pursuing degree** | | |
| Bachelor's degree | 19 | 51 |
| Master's degree | 18 | 49 |
| **Years since pursuing the degree** | | |
| Less than 1 year | 18 | 49 |
| Less than 2 years | 7 | 19 |
| Less than 3 years | 3 | 8 |
| Less than 4 years | 6 | 16 |
| 4 years or more | 3 | 8 |
| **Computer access** | | |
| Have access to a PC or laptop | 37 | 100 |
| **Internet access** | | |
| Reliable high-speed Internet access | 36 | 97 |
| Unreliable or slow Internet access | 1 | 3 |

## 4.2 Perceptions of Educational Chatbots in Higher Education

The first five questions in both surveys explored respondents' perceptions of educational chatbots. In the student survey, the first question asked how future chatbots could better meet learners' needs. The second focused on tasks where students preferred chatbot interaction over human support. The third asked whether chatbots can offer personalized learning experience. These same questions were included in the staff survey but framed from a teaching perspective, allowing for comparison between student and staff views. The remaining two questions assessed the importance of two chatbot key features: maintaining conversational context and customization to accommodate individual preferences and learning styles. Both features have become increasingly feasible due to GenAI advancements.

### 4.2.1 Staff's Perceptions

A substantial majority of staff respondents (87%, n=61) envisioned that future educational chatbots could provide learners with 24/7 learning support, while slightly over half (54%) anticipated personalized learning paths. For specific tasks, most staff respondents preferred that chatbots assist students in searching for papers or books (66%) and answering general course-related questions (64%). When asked about chatbots' capability to provide personalized learning experiences, 54% believed chatbots could somehow offer such experiences, 23% believed they could significantly do so, while 7% were unsure and 16% did not believe that chatbots could be effective in this regard. Lastly, 61% of staff respondents considered maintaining conversational context extremely important, whereas only 21% viewed customization as equally important.

### 4.2.2 Students' Perceptions

A vast majority of student respondents (86%, n=37) anticipated receiving 24/7 learning support from future chatbots. Additionally, more than half envisioned that chatbots could engage learners through interactive experiences (57%), provide personalized learning paths (57%), and offer adaptive assessment and feedback based on individual performance (54%). In terms of preferred chatbot assistance, most students wanted help with searching for papers and books (65%) and finding resources for assignments or projects (51%). When asked about chatbots' capability to provide personalized learning experiences, 70% believed chatbots could somehow offer such experiences, 14% believed they could significantly do so, while 5% were unsure and 11% did not believe chatbots could be effective in this regard. Lastly, 70% of student respondents considered maintaining conversational context extremely important, whereas only 22% viewed customization as equally important.

### 4.2.3 Comparison and Additional Insights

The majority of both groups envisioned that future chatbots will offer 24/7 learning support and personalized learning paths. Additionally, most student respondents expected chatbots to engage learners through interactive experiences (57%, n=37) and provide adaptive assessment and feedback (54%). While no unique comments were found among students, staff respondents shared diverse perspectives. Three envisioned chatbots that could perform teaching and assessment entirely, create digital teaching content, or encourage and guide users to seek human help. In contrast, one staff respondent strongly opposed chatbot use, expressing concern that it would reduce teaching resources and quality. Table3 presents the comparison of envisioned chatbot capabilities between the groups.

*Table 3 Comparison of Envisioned Chatbot Capabilities*

| Staff (n=61) | Students (n=37) |
| --- | --- |
| **From the teaching perspective, how do you envision chatbots evolving in educational platforms in the future to better meet the needs of learners?** | **How do you envision chatbots evolving in educational platforms in the future to better meet the needs of learners?** |
| 24/7 Learning Support (87%) | 24/7 Learning Support (86%) |
| Personalized Learning Paths (54%) | Interactive Learning Experiences & Personalized Learning Paths (57%) |
| Interactive Learning Experiences (49%) | Adaptive Assessment and Feedback (54%) |
| Adaptive Assessment and Feedback (44%) | Career Guidance and Mentorship (38%) |
| Career Guidance and Mentorship (41%) | *No Response (3%)* |
| Other (5%): chatbot teaching & assessment; digital teaching content creation; supporting & guiding the user to seek human help | |
| Other (opposed chatbots) (2%) | |

Regarding chatbot assistance, helping students search for papers or books was the most preferred task among both groups. A majority of student respondents (51%, n=37) and a considerable portion of staff (44%, n=61) also preferred chatbots for finding resources for assignments or projects. However, while most staff (64%, n=61) supported delegating general course-related questions to chatbots, only 22% of student respondents (n=37) selected this option. Overall, staff respondents appeared more open to using chatbots for various support tasks than students (see Table 4).

*Table 4 Comparison of Preferred chatbot assistance*

| Staff (n=61) | Students (n=37) |
| --- | --- |
| **From the teaching perspective, would you prefer your students interacting with a chatbot for certain tasks over human instructors or support staff? If yes, please specify some tasks.** | **Would you prefer interacting with a chatbot for certain tasks over human instructors or support staff? If yes, please specify some tasks.** |
| Searching for papers or books (66%) | Searching for papers or books (65%) |
| Asking general questions about the course (64%) | Finding resources for assignments or projects (51%) |
| Renting a book (48%) | Accessing course materials (32%) |
| Finding resources for assignments or | Renting a book (30%) |

| Staff (n=61) | Students (n=37) |
|---|---|
| projects (44%) | |
| Accessing course materials (41%) | Asking general questions about the course & Getting updates on course announcement (22%) |
| Getting updates on course announcement (36%) | Scheduling appointments or meetings (19%) |
| Scheduling appointments or meetings (28%) | Asking for feedback on assignments or projects (14%) |
| Provide students with advice on personal study plan (20%) | Personal study plan (5%) |
| Asking for feedback on assignments or projects (16%) | Other tasks (3%): Concept clarification |
| Other tasks (2%): Assist students with special education needs | *No response (16%)* |
| Other (opposed chatbots) (2%) | |
| *No response (7%)* | |

It is worth noting that, compared to the high response rate to the question on envisioned chatbot capabilities, 16% of students (n=37) and 7% of staff (n=61) skipped the question on preferred chatbot assistance. This may suggest that some respondents in both groups prefer human assistance over chatbot support. To conclude, while most respondents anticipated that future chatbots could offer 24/7 support and personalized learning—and considered maintaining conversational context to be extremely important—attention should also be given to the concerns of those who prefer human interaction. Addressing these preferences will be essential when designing and implementing chatbot-based support in higher education.

## 4.3 Usage of Chatbots and Other GenAI Applications in Higher Education

Staff respondents were asked how often they had used chatbots or other GenAI applications for work in the past six months, with the options: "never," "less than monthly," "monthly," "weekly," and "daily." Those who selected "never" received a follow-up, multi-select question to specify reasons for not using GenAI at work. Respondents who had used GenAI were instead asked to select activities that best described their usage. Student respondents were asked similar questions about their use of GenAI for study, following the same structure. All follow-up questions included an "other" option with a text field for elaboration.

### 4.3.1 Staff's Usage

By the time of the survey, the vast majority of staff respondents (90%, n=61) had used GenAI for work to varying degrees: 18% less than monthly, 15% monthly, 31% weekly, and 26% daily. Among those who had used GenAI, the top three activities were *clarifying concepts and terminologies* (62%, n=55), *brainstorming* (58%, n=55), and *grammar checking* (56%, n=55). Six respondents reported not using GenAI at all, reporting *lack of interest* (67%, n=6) and concerns about *research integrity* (50%, n=6) as their main reasons.

### 4.3.2 Students' Usage

Similarly, 89% of student respondents (n=37) had used GenAI for their studies to varying degrees: 21% less than monthly, 13% monthly, 47% weekly, and 8% daily. Among those who had used GenAI, the top three activities were *clarifying concepts and terminologies* (82%, n=33), *brainstorming* (52%, n=33), and *debugging software code* (36%, n=33). Four respondents had not used GenAI for study, reporting *privacy* (75%, n=4), *study integrity* (75%, n=4), and dissatisfaction with the *quality of generated content* (75%, n=4) as their main concerns.

### 4.3.3 Comparison and Additional Insights

The results showed that the vast majority of both staff and student respondents had used GenAI for work or study. Notably, daily use was more common among staff (26%, n=61) than students (8%, n=37). While both groups primarily used GenAI for concept clarification and brainstorming, staff reported a broader range of activities (see Table 5). A majority of staff respondents also used GenAI for *grammar checking* (56%, n=55), *summarizing* (55%), and *translation* (53%). Nearly half used it for *generating images* (47%), and about one-third for *debugging software code* (31%). Additionally, 15% reported other uses such as text generation and improvement, analyzing articles, code generation, and providing feedback. A few had also used GenAI to generate audio (4%) and video (2%). Compared to staff, fewer students reported using GenAI for these activities—except for debugging software code (36%, n=33). No student respondents had used GenAI to generate audio or video for their studies.

*Table 5 Comparison of GenAI Usage*

| Staff (n=55) | Students (n=33) |
| --- | --- |
| **Please select one or more activities that best describe your usage of chatbots or other generative AI applications for work-related tasks.** | **Please select one or more activities that best describe your usage of generative AI applications for study related tasks.** |
| Clarifying concepts and terminologies (62%) | Clarifying concepts and terminologies (82%) |
| Brainstorming (58%) | Brainstorming (52%) |
| Grammar checking (56%) | Debugging software code (36%) |
| Summarizing (55%) | Summarizing (33%) |
| Translation (53%) | Grammar checking (21%) |

| Staff (n=55) | Students (n=33) |
|---|---|
| Generating images (47%) | Translation (15%) |
| Debugging software code (31%) | Other (9%): code generation, text improvement, creating tables from text |
| Other (15%): text or code generation, finding themes from articles, text improvement, providing feedback | Generating images (6%) |
| Generating audio (4%) | *Generating audio or video (0%)* |
| Generating video (2%) | |

## 4.4 Impacts of GenAI on Staff and Students

Staff respondents were asked how much chatbots or other GenAI applications have changed the way they work, using a Likert scale with the following options: "not at all," "a little," "moderately," "noticeably," and "very much." Those who selected any option other than "not at all" received a follow-up question asking for both positive and negative examples of how the technology has changed their work. Similar questions were asked of student respondents, with the context shifted to their studies. To capture broader impacts, these questions were also presented to respondents who reported not having used GenAI in the past six months.

### 4.4.1 Impacts on Staff

A large majority of staff respondents (80%, n=61) felt that GenAI had changed the way they work to various extents. Specifically, 18% reported a little change, 31% a moderate change, 20% a noticeable change, and 11% said GenAI had changed their work very much. Among those who reported changes, only one had not used GenAI for work.

After rating perceived changes, 34 respondents shared further experiences. The most common theme was a self-reported *productivity boost*, mentioned by 30 respondents. GenAI was often credited with improving efficiency in *information search* (11 respondents), *writing* (8), *translation* (5), and *programming tasks* (6). Respondents also described how GenAI helped streamline *research* (5), *teaching* (3), and *administrative tasks* (2). In research, GenAI had been used across various stages, including identifying gaps and errors (1), transcribing interviews (1), streamlining data acquisition and analysis (1), selecting statistical methods (1), and obtaining second opinions (1). In teaching, it was used to assess lecture coverage (1), generate exercise ideas (1), design exams (1), and manage workload (1). For administrative work, one respondent used GenAI to draft impact statements for grants, and another to create spreadsheet functions.

Despite the reported productivity gains, seven respondents also expressed a critical stance toward GenAI, emphasizing the need to *think critically* (6) and *verify its responses* (3). Additional challenges and concerns were raised by ten respondents. Six described *the shortcomings of current GenAI tools* and how they coped with them. These included disappointing summarization (2), shallow responses (1), the need to revise generated text (1), correcting code errors (1), and a preference for domain-specific tools (1). Four respondents raised *concerns about students' use of GenAI*. One, who had not used GenAI, remarked, "...despite the instructions, the use of artificial intelligence has not been made explicit, giving the impression that the text was written by the students themselves." Another respondent reported having to modify essay assignments as a result. Other individual concerns included *protecting intellectual property in research* and *the risk of reduced critical thinking* due to over-reliance on GenAI.

### 4.4.2 Impacts on Students

Among student respondents, most (64%, n=37) felt that GenAI had changed the way they study to varying extents. Specifically, 32% reported a little change, 19% a moderate change, 8% a noticeable change, and 5% said GenAI had changed their study habits very much. All of these respondents had used GenAI for study purposes.

18 student respondents shared examples of how GenAI had impacted their studies. All reported *productivity boosts* across various activities, including *information search* (9 respondents), *writing* (5), *programming* (3), brainstorming (1), receiving feedback (1), and translation (1). Notably, three respondents mentioned using GenAI more frequently than traditional search engines for information searches.

While all respondents to this question reported productivity gains, five also expressed a *critical mindset toward GenAI*, with two specifically mentioning *the need to verify its responses*. Four respondents raised *challenges and concerns*. Two pointed out shortcomings such as excessive code changes during reviews and translation errors and inconsistencies. One noted that online exams had become more difficult, and another voiced concern about GenAI's effect on creativity, stating, "I think this could also be impacting my creativity as I don't need to think about what I'm going to do very long."

### 4.4.3 Comparison and Additional Insights

Overall, both staff and student respondents reported changes in their work and study due to GenAI, with staff more frequently perceiving moderate to significant impacts. Responses to the open-ended question provided additional insights into these changes (see Table 6). Thematic analysis revealed productivity boosts in information search, writing, and programming across both groups. Some staff also reported using GenAI to streamline teaching, research, and administrative tasks. Beyond productivity gains, both groups emphasized the importance of adopting a critical mindset toward GenAI responses and coping with its shortcomings. Group-specific concerns also emerged: staff mentioned students' use of GenAI, intellectual property protection, and the risk of reduced critical thinking, while students highlighted challenges such as more difficult online exams and the potential impact on creativity. Overall, the findings suggest that some respondents may be experiencing a dilemma between the productivity benefits of GenAI and its possible side effects.

*Table 6 Comparison of GenAI Impacts*

| Staff (34 respondents) | Students (18 respondents) |
|---|---|
| **Self-reported productivity boost** (staff: 30, students: 18) | |
| Improving efficiency in information search (11) | Improving efficiency in information search (9) |
| Increasing writing productivity (8) | Increasing writing productivity (5) |
| Streamlining programming tasks (6) | Streamlining programming tasks (3) |
| Speeding up translation (5) | Other (4): translation, brainstorming, getting feedback, learning in general |
| Research (5): identifying gaps and errors, looking for statistical methods, transcribing interviews, data acquisition and analysis, getting second opinions from GenAI | |
| Teaching (3): assessing lecture material coverage, getting ideas for exercises, exam design, coping with increased workload | |
| Administration (2): making impact statements for grants, creating spreadsheet functions | |
| Other (7): brainstorming, creating images and large amounts of content, helping in routines, repetitive and menial tasks, enhancing skill development, improving efficiency in general | |
| **Adopting a critical mindset towards GenAI responses** (staff: 7, students: 5) | |
| Need to be critical (6) | Need to be critical (4) |
| Need to verify GenAI responses (3) | Need to verify GenAI responses (2) |
| **Emerging challenges and concerns** (staff: 10, students: 4) | |
| Coping with GenAI shortcomings (6) | Coping with GenAI shortcomings (2) |
| Concerns on students' use of GenAI (4): inappropriate use of GenAI in writing assignments, motivation letters, and emails | Online exams have become tougher (1) |
| Concern about protecting intellectual property in research (1) | Risk of reduced creativity (1) |
| Risk of reduced critical thinking (1) | |

## 4.5 Concerns About Using Chatbots and GenAI Applications in Higher Education

This section was identical in both surveys and included nine Likert-scale items and one open-ended question. The Likert items presented statements expressing concerns about using the technology in higher education across the following aspects: 1) academic integrity, 2) creativity, 3) critical thinking and communication skills, 4) privacy and data ownership, 5) the quality of GenAI responses, 6) security, 7) social interaction, 8) GenAI dependency, and 9) copyright infringements. Respondents rated each statement on a five-point scale, from "I am not concerned at all" (1) to "I am very concerned" (5). After the ratings, they were invited to list any additional major concerns in an open-ended question.

### 4.5.1 Staff's Concerns

The Likert-scale results showed that most staff respondents were very concerned about *the quality of GenAI responses* (51%, n = 61), including issues such as incorrect, outdated, biased, or inappropriate outputs. *Privacy and data ownership* followed, with 43% expressing strong concern about how personal data is used, stored, and owned when using GenAI. Around one-third of respondents were very concerned about the impact of GenAI on *academic integrity* (33%) and the risk of *copyright infringements* (33%). These were the top three areas of concern. See Table 7 for the full distribution of responses.

*Table 7 Comparison of Concern Ratings (Staff Respondents % / Student Respondents %)*

| Aspect | Level of Concerned (%) | | | | |
|---|---|---|---|---|---|
| | **Not (1)** | **Little (2)** | **Moderate (3)** | **Quite (4)** | **Very (5)** |
| Academic integrity | 5 / 5 | 10 / 14 | 18 / 19 | 33 / 32 | 33 / 30 |
| Creativity | 41 / 19 | 20 / 14 | 15 / 22 | 16 / 22 | 7 / 22 |
| Critical thinking & communication skills | 15 / 8 | 8 / 11 | 11 / 16 | 30 / 32 | 31 / 32 |
| Privacy & data ownership | 8 / 3 | 7 / 11 | 13 / 19 | 26 / 22 | 43 / 46 |
| The quality of GenAI responses | 3 / 0 | 8 / 3 | 15 / 5 | 20 / 30 | 51 / 62 |
| Security | 8 / 11 | 16 / 16 | 18 / 14 | 26 / 30 | 28 / 30 |
| Social interaction | 49 / 32 | 23 / 24 | 11 / 11 | 13 / 8 | 2 / 19 |
| GenAI dependency | 43 / 22 | 21 / 14 | 10 / 14 | 16 / 30 | 5 / 22 |
| Copyright infringements | 8 / 8 | 18 / 16 | 16 / 19 | 18 / 27 | 33 / 30 |

*Staff (n=61), students (n=37)*

Additionally, 15 staff members shared concerns in the open-ended question. The most common theme was the risk of *impeding students' learning*, mentioned by 11 respondents. These responses indicated that GenAI had been used not only for completing essays and programming assignments but also for cheating. Some staff expressed concern that students' use of GenAI had rendered *certain assessment methods, such as essay writing, inadequate* (3). Others noted that over-reliance on GenAI could hinder the development of *higher-order thinking skills* (6), ultimately affecting student learning. See Table 8 for a summary of the identified themes.

*Table 8 Comparison of Recurrent and Additional Concern Themes*

| Staff (15 respondents) | Students (5 respondents) |
|---|---|
| **Reiterated concerns from Likert-scale items** | |
| Quality of GenAI responses (6) | Quality of GenAI responses (4) |
| Critical thinking (6) | Creativity (1) |
| Academic integrity (3) | Social interaction (teacher-student) (1) |
| Over-dependence (3) | |
| Social interaction (teacher-student) (1) | |
| Copyright infringements (1) | |
| **Additional concerns** | |
| Impeding students' learning (11) | Impeding students' learning (5) |
| Impeding general higher-order thinking besides critical thinking (6) | Increasing energy consumption (1) |
| Inadequate assessments (3) | Increasing social inequality (1) |
| Criminal use (1) | |
| General behavioral impacts on users (1) | |
| Reducing interest in life-long learning (1) | |
| Risk of neglecting stakeholder concerns when adopting a new technology (1) | |
| Suboptimal personalization (1) | |
| Unknown negative impacts on research universities (1) | |
| Unspecified long-term consequences (1) | |

### 4.5.2 Students' Concerns

The majority of student respondents were also very concerned about *the quality of GenAI responses* (62%, n = 37). The next most common concerns were *privacy and data ownership* (46%), followed by the potential negative impact on *critical thinking and communication skills* (32%). Five students responded to the open-ended question, and all expressed a similar concern: that using GenAI could hinder their learning.

### 4.5.3 Comparison and Additional Insights

While staff and student respondents rated many aspects similarly, some differences emerged (see Table 7). Since the Likert-scale questions in this section are identical for staff and students, the Mann–Whitney U test was applied. The tests revealed significant differences in concerns about the negative impact on individual creativity (p = 0.004) and the risk of becoming overly dependent on GenAI technology (p = 0.001). Student respondents were more concerned than staff about losing creativity and becoming overly reliant on GenAI.

Recurrent and additional concerns emerged from the open-ended responses (see Table 8). Notably, the potential of GenAI to *impede students' learning* stood out as a key concern besides those identified in the Likert-scale items. The responses also revealed unique concerns extending beyond the context of higher education, particularly environmental and social impacts. For example, one student wrote: "Generative artificial intelligence is far too heavy in terms of resource requirements, causing an increase in energy consumption when it should be radically reduced in society." In addition to this section, we also identified concerns in other parts of the surveys, which will be further summarized and discussed in the discussion section.

## 4.6 Ideation of GenAI Applications in Higher Education

This section of the surveys explored relevant user stories, use cases, and functional requirements for applying GenAI in higher education from the perspectives of staff and students. As a warm-up, respondents first answered single-select questions on the perceived usefulness of selected GenAI applications. They were then invited to share additional use cases, user stories, or functional requirements they found very useful through an open-ended question.

### 4.6.1 Staff's Perspective

Four user stories, targeting lecturers, researchers, project managers, and research assistants, asked staff respondents to rate their usefulness on a scale from "not useful at all" to "very useful." A majority (56%, n = 61) found the user story of a researcher using GenAI to identify research gaps from a literature review to be quite or very useful. Similarly, about 53% found the story where a research assistant uses GenAI for programming tasks to be quite or very useful. In contrast, fewer respondents considered the user stories about GenAI assisting in drafting funding applications (46%) and assessing student assignments (41%) to be quite or very useful. Table 9 shows the distributions of perceived usefulness across the presented user stories.

*Table 9 Staff Response Distribution on Perceived Usefulness of Selected User Stories*

| User Story | Perceived Usefulness (%, n=61) | | | | |
|---|---|---|---|---|---|
| | Not at all (1) | Little (2) | Moderately (3) | Quite (4) | Very (5) |
| As a researcher, I want to use the application to help me review literature, so I can identify research gaps more effectively | | | | | |
| | 13 | 15 | 11 | 26 | 30 |
| As a research assistant, I want to use the application to help me in programming, so I can develop software prototypes effectively | | | | | |

| User Story | Perceived Usefulness (%, n=61) | | | | |
|---|---|---|---|---|---|
| | 11 | 5 | 23 | 30 | 23 |
| As a project manager, I want to use the application to help me draft funding applications, so I can improve the quality of my applications | | | | | |
| | 15 | 7 | 26 | 13 | 33 |
| As a lecturer, I want to use the application to help me assess students' assignments, so I can make assessment more effectively | | | | | |
| | 20 | 20 | 16 | 18 | 23 |

After rating the provided user stories, 22 staff respondents answered the subsequent open-ended question. A thematic analysis was conducted to primarily elicit high-level requirements from the responses. While 20 respondents shared use cases, user stories, or requirements they found very useful, four also expressed skepticism or concerns.

By grouping similar or related use cases, user stories, and requirements, we identified high-level requirements relevant to three core functions commonly found in research universities: teaching, research, and administration. For teaching, GenAI could be used to *create engaging content* and *streamline course assessment*, from design to implementation. In research, GenAI may support not only *literature reviews* but also *research design* (such as identifying suitable methods) and accelerate or automate *data analysis*, especially for qualitative data, including transcribing and translating interviews and performing thematic and summative analyses of interview or survey responses. For administration, GenAI has the potential to *streamline research and funding application processes* and *support international student admissions*. In addition, we identified cross-functional requirements relevant across all three areas. GenAI could assist in efficiently *producing professional documents*: helping users overcome writer's block, offering feedback, and refining texts. Actors across functions could also benefit from harnessing and customizing advanced AI for *personalized assistance*. Finally, five relevant non-functional requirements emerged: *consistent customization, simplified prompt engineering, large document input capacity, human-like interaction, and market-comparable quality*. Table 10 summarizes the identified high-level functional and non-functional requirements.

*Table 10 High-Level Requirements Identified from Staff's Open-Ended Responses (22 Respondents)*

| Functional requirement (number of respondents) | Initial labels from the responses |
|---|---|
| **For teaching** (5) | |
| *Create engaging teaching content* (1) | Create digital teaching content (1) |
| *Streamline course assessment* (4) | As a lecturer, I want to use the application to help me write assignments, so I can offer students more variety (1), Create and assess |

| Functional requirement (number of respondents) | Initial labels from the responses |
|---|---|
| | personalized assignments (1), Assess student assignments (2), Assess student competence (1) |
| **For research** (6) | |
| *Streamline literature review* (4) | Do literature review (1), Get Scopus AI-like services (1), Summarize large amount of research papers (1), The application should provide accurate and relevant references (1) |
| *Support research design* (1) | Find suitable research methods (1) |
| *Streamline qualitative analysis* (1) | Transcribe and translate interviews (1), Do thematic and summative analysis of interviews and survey responses (1) |
| *Automate data analysis* (1) | Automate data analysis (1) |
| **For administration** (5) | |
| *Streamline research and funding application processes* (4) | Write Excel functions (1), Fill template from text (1), Automate budgeting and resource planning in research applications (1), Evaluate funding applications against funding call criteria (1) |
| *Support international student admissions process* (1) | The AI chatbot should support the international student admissions process (1) |
| **Cross-functional** (9) | |
| *Create professional documents efficiently* (7) | Generate text snippets to overcome the writer's block (1), Generate images (1), Generate marketing materials (1), Create and translate project materials (1), Create presentation drafts from information input (1), Customize presentations (1), Check English (1), Get feedback for different kinds of documents (1) |
| *Harness and customize advanced AI for personalized assistance* (3) | The application should provide access to advanced models (1), As a user, I want to easily customize AI so that I can use it in different tasks (1), Get personal assistance (1) |
| **Non-functional requirements** (2) | |
| Consistent customization (1), Simplified prompt engineering (1), Large document input capacity (1), Human-like interaction (1), Market-comparable quality (1) | |

Regarding skepticism and concerns, we identified two main categories: concerns about GenAI itself and concerns about organizational support for technology. In terms of GenAI, one respondent remarked, "Until LLMs are trustworthy, they cannot be useful." Another noted that GenAI can still produce inaccurate responses, which could have serious consequences if left unchecked. This respondent also emphasized that GenAI cannot be reliably used for assessing student assignments and should not replace essential interactions between students and teachers, ultimately expressing strong disagreement with the use of GenAI in higher education. In terms of organizational support, one respondent expressed doubt about the quality of GenAI applications provided by the university, while another lamented the lack of training and resources available through the faculty.

## 4.6.2 Students' Perspective

Student respondents were first asked how beneficial a GenAI personal study coach would be in a challenging course. A vast majority considered such an application beneficial to varying degrees: 22% (n = 37) found it somewhat beneficial, 29% moderately beneficial, 27% very beneficial, and 14% extremely beneficial. In contrast, 8% felt it was not beneficial.

Following that, students were asked to rate the usefulness of eight specific user stories, on a scale from "not useful at all" to "very useful". A majority indicated that programming assistance (60%, n = 37) and finding literature relevant to their studies (54%) were particularly useful, rating these user stories as either "quite useful" or "very useful." Additionally, nearly half of the respondents rated the following user stories as quite or very useful: brainstorming thesis topics (49%), receiving math tutoring or assistance (49%), and getting tips and feedback on job applications (49%). Furthermore, 46% found using GenAI to help them understand and excel in course-specific topics to be quite or very useful. In contrast, only 30% of respondents considered GenAI useful for managing their projects, and just 22% rated its usefulness for managing personal study plans as quite or very useful. Table 11 shows the distribution of student responses for each user story.

*Table 11 Student Response Distribution on Perceived Usefulness of Selected User Stories*

| User Story | Perceived Usefulness (%, n=37) | | | | |
|---|---|---|---|---|---|
| | Not at all (1) | Little (2) | Moderately (3) | Quite (4) | Very (5) |
| As a user, I want to use the application to receive programming assistance or coding support, so I can enhance my coding skills more effectively | | | | | |
| | 14 | 11 | 16 | 19 | 41 |
| As a user, I want to use the application to find relevant literature for my course topic from institutional databases, so I can enhance my understanding and research skills | | | | | |
| | 14 | 5 | 27 | 27 | 27 |
| As a user, I want to use the application to help me brainstorm on thesis topic, so I can plan my thesis work more effectively | | | | | |
| | 19 | 11 | 22 | 27 | 22 |
| As a user, I want to use the application to provide me with tips and feedback on my job applications, so I can improve the quality and attractiveness of my job applications | | | | | |
| | 11 | 24 | 11 | 19 | 30 |
| As a user, I want to use the application to receive tutoring or assistance in math, so I can improve my math skills more effectively | | | | | |
| | 16 | 16 | 19 | 30 | 19 |
| As a user, I want to use the application to receive help with course-specific topics (e.g., wireless communication, data analysis), so I can better understand and excel in my studies | | | | | |

| User Story | Perceived Usefulness (%, n=37) | | | | |
|---|---|---|---|---|---|
| | 11 | 11 | 32 | 19 | 27 |
| As a user, I want to use the application to receive assistance in project planning or execution, so I can effectively manage and complete my projects | | | | | |
| | 24 | 19 | 27 | 22 | 8 |
| As a user, I want to use the application to help me plan and update Personal Study Plan, so I can manage my studies more effectively | | | | | |
| | 38 | 27 | 11 | 8 | 14 |

In contrast to staff respondents, only two students answered the open-ended question, and their responses did not reveal additional use cases. One respondent considered programming guidance and support to be very useful, while the other felt it would be helpful for clarifying unclear matters and answering questions outside of lectures.

### 4.6.3 Key Takeaway

Among the selected user stories, the majority of staff respondents found using GenAI for literature reviews and programming tasks especially useful. For students, most felt that having a personal study coach could be beneficial to varying extents. They also considered the user stories involving programming assistance and finding relevant literature for their courses to be particularly useful.

Furthermore, the thematic analysis of open-ended responses suggested that similar or related ideas could be grouped to develop high-level requirements aimed at better supporting key organizational processes. Although no high-level requirements were identified from student responses due to the limited number of open-ended replies, the analysis of staff responses revealed relevant functional and non-functional requirements that could be further elaborated and validated in collaboration with staff members.

# 5 Bridging Stakeholder Needs with GenAI Capabilities in Education

The staff and student surveys offered insights into concrete user stories, functional needs, and expectations regarding GenAI applications in higher education. The responses uncovered a wide spectrum of perceived usefulness and practical opportunities, spanning teaching, research, and administrative domains. Notably, staff valued GenAI's potential to assist in literature reviews and programming tasks, while students saw promise in personalized study support, programming help, and research guidance. The thematic analysis of staff feedback led to the identification of high-level functional and non-functional requirements, such as support for creating engaging teaching content, streamlining course assessments, automating data analysis, and facilitating the drafting of professional documents. These requirements underscore the demand for cross-functional, intelligent systems capable of personalization, human-like interaction, and high adaptability. Moreover, concerns from some respondents revealed the need for transparent, trustworthy, and institutionally supported GenAI implementations.

In light of these findings, a representative GenAI-powered educational platform is proposed to translate these needs into a comprehensive, scalable solution. The platform framework integrates various AI-driven tools and services; including chatbots, personalized learning paths, analytics dashboards, and adaptive assessment systems; designed to enhance teaching effectiveness, streamline administrative workflows, and support personalized student learning journeys. The following section presents this conceptual framework in detail, illustrating how emerging technologies, robust architecture, and modular components can be orchestrated to meet the diverse functional and organizational requirements identified from the surveys.

## 5.1 Conceptualizing a GenAI-Driven Ecosystem

The diagram in Fig. 2 represents an AI-powered educational platform that integrates various digital tools and services to enhance learning experiences for students, educators, and administrators. The platform leverages AI to create personalized, efficient, and data-driven educational processes, supported by cloud services for scalability and security. The basic technical details are as follows:

- **Key Stakeholders**:

    - Students: Students access the platform to engage in learning activities through AI-powered tools such as the Online Library, the Learning Management System (LMS), and Personalized Learning Paths. The students are also allowed to chat with the bot with restricted access to certain components of platform.

    - Administrators: Adminstractors interact with the platform to manage and optimize educational processes, including tracking student progress and managing content through the Analytics Dashboard and the LMS. The administrators will also have full access via chatbot.

    - Educators: Educators use the platform to design curriculum, track student performance, and publish research. They can utilize the AI-Powered Chatbot for assistance and generate personalized content for students.

- **Core Components**:

    - AI-Powered Chatbot: The chatbot functions as a central system here. It interacts with students, educators, and administrators to assist with queries, provide information, and guide users through the platform's features. The chatbot is capable of integrating with various modules, such as

the LMS and Analytics Dashboard, and can respond to student inquiries or help educators and administrators with tasks.

o Online Library Access: Students and educators can retrieve learning resources and digital content from an extensive online library. The library is integrated with the cloud for scalable data storage and retrieval.

o Learning Management System (LMS): The LMS handles the management of educational content, course administration, student assessments, and collaboration between students and instructors. AI enhances LMS functionality by providing real-time recommendations, adaptive learning suggestions, and personalized content.

o Analytics Dashboard: Administrators and educators can view key performance indicators and student progress. The dashboard uses AI-driven analytics to offer insight on student participation, learning outcomes, and areas for improvement.

o Personalized Learning Paths: AI algorithms assess each student's learning style and progress to create customized learning pathways. This adaptive feature ensures that learning is tailored to individual needs, offering personalized content and assessments. AI can enhance assessments by automatically generating adaptive assessments based on the student's skill level. This allows the platform to create customized quizzes and exams that change based on user progress or weaknesses. Adaptive learning allows assessments to become more dynamic and student-focused, ensuring that each test is tailored to the student's current knowledge and ability. AI can monitor performance and generate questions of varying difficulty.

o Publication Platform: This platform enables educators to streamline their academic writing processes, manage research findings, and organize educational resources. It integrates AI functionalities for efficient document formatting, precise citation management, and robust plagiarism detection. Furthermore, previously published papers are indexed chronologically to facilitate rapid information retrieval and serve as a comprehensive historical record.

- **Cloud Services and Integrations**:

  o Single Sign-On (SSO): The platform offers secure and seamless user access through Single Sign-On (SSO) integration. This allows users to authenticate once and gain continuous access to all services without repeated logins. Comprehensive identity and access management (IAM) underpins this functionality, providing advanced role-based access control (RBAC) to ensure fine-grained authorization based on user roles and specific use cases.

- Multilingual Support: The platform supports multilingual content and interactions. AI-powered translation services ensure students and educators can access content in their preferred language.

- Automated Testing and Assessments: AI assists in creating and grading tests and assessments. It offers automated feedback and insights, helping students and educators improve their learning strategies.

- Cloud Infrastructure: All data, content, and services are hosted on Cloud Services, ensuring scalability, availability, and security. The platform can expand to accommodate more users and resources while maintaining high performance.

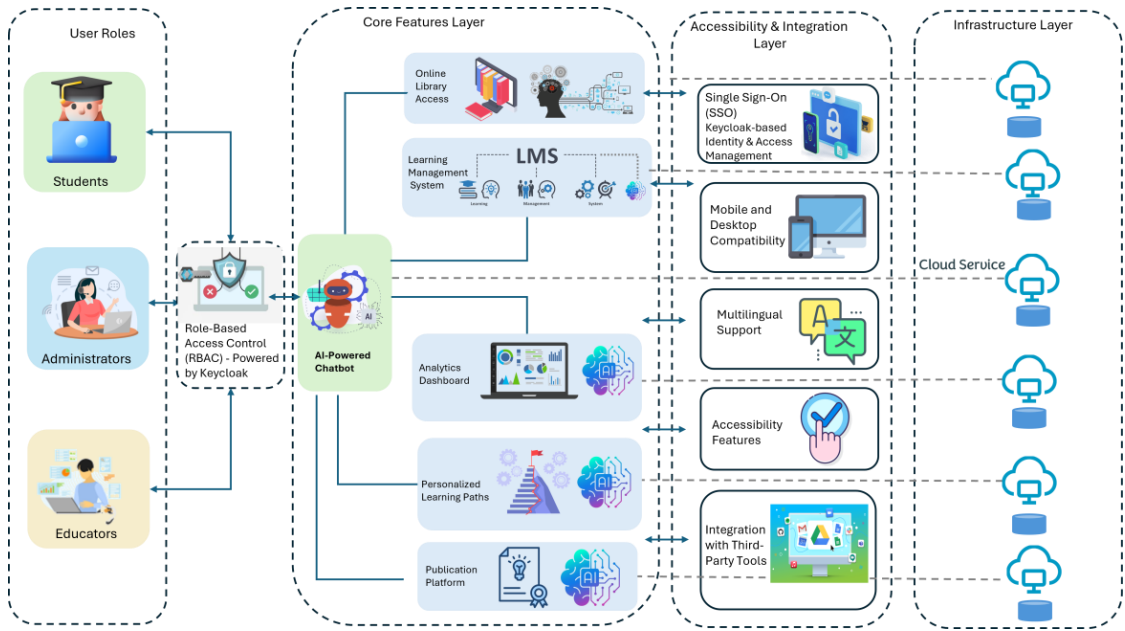

*Fig. 2 AI-Powered Educational Platform*

A proposed technology stack for the GenAI powered eductaional pltaforms is as shown in Fig. 3. This diagram visually organizes the layers of technology involved in developing an AI-driven educational system. This organized stack provides a comprehensive view of the technologies required to build and run a generative AI-powered educational platform.

- UI Layer: This layer enables seamless interaction for students, educators, and administrators through intuitive web and mobile interfaces. It includes dashboards for analytics, AI-powered chatbots for support, and interactive tools for personalized learning. Technologies like React, Angular, Vue.js, and Flutter are used to build these interfaces.

- Application Layer: This serves as the system's core, managing resource sharing, personalization, and content delivery. It powers learning management, AI-driven learning path optimization, and NLP-based content summarization using technologies like LangChain, LlamaIndex, Haystack, and Semantic Kernel.

- API Layer: This layer facilitates seamless integration with external services, enabling authentication via Single Sign-On (SSO), access to third-party tools like library systems, and multilingual support through translation APIs. Key technologies include OpenID Connect, Google Translate API, and RESTful APIs for LMS integration.

- LLM Layer: The LLM Layer drives NLP, enabling content generation, chatbot intelligence, and contextual recommendations. It utilizes technologies like GPT-4 for conversational AI, T5 for summarization, BERT for query comprehension, and LLaMA for domain-specific language tasks.

- Data Layer: The Data Layer handles the storage, organization, and retrieval of structured and unstructured data, including user profiles, course materials, and analytics. It employs technologies like PostgreSQL for databases, FAISS for semantic search, and Neo4j or Milvus for managing knowledge graphs.

- Infrastructure Layer: The Infrastructure Layer ensures scalability, reliability, and secure operations by leveraging cloud-based resources for hosting, storage, and compute needs. It uses platforms like AWS, Google Cloud, and Microsoft Azure for efficient application deployment, data backups, and elastic AI model processing.

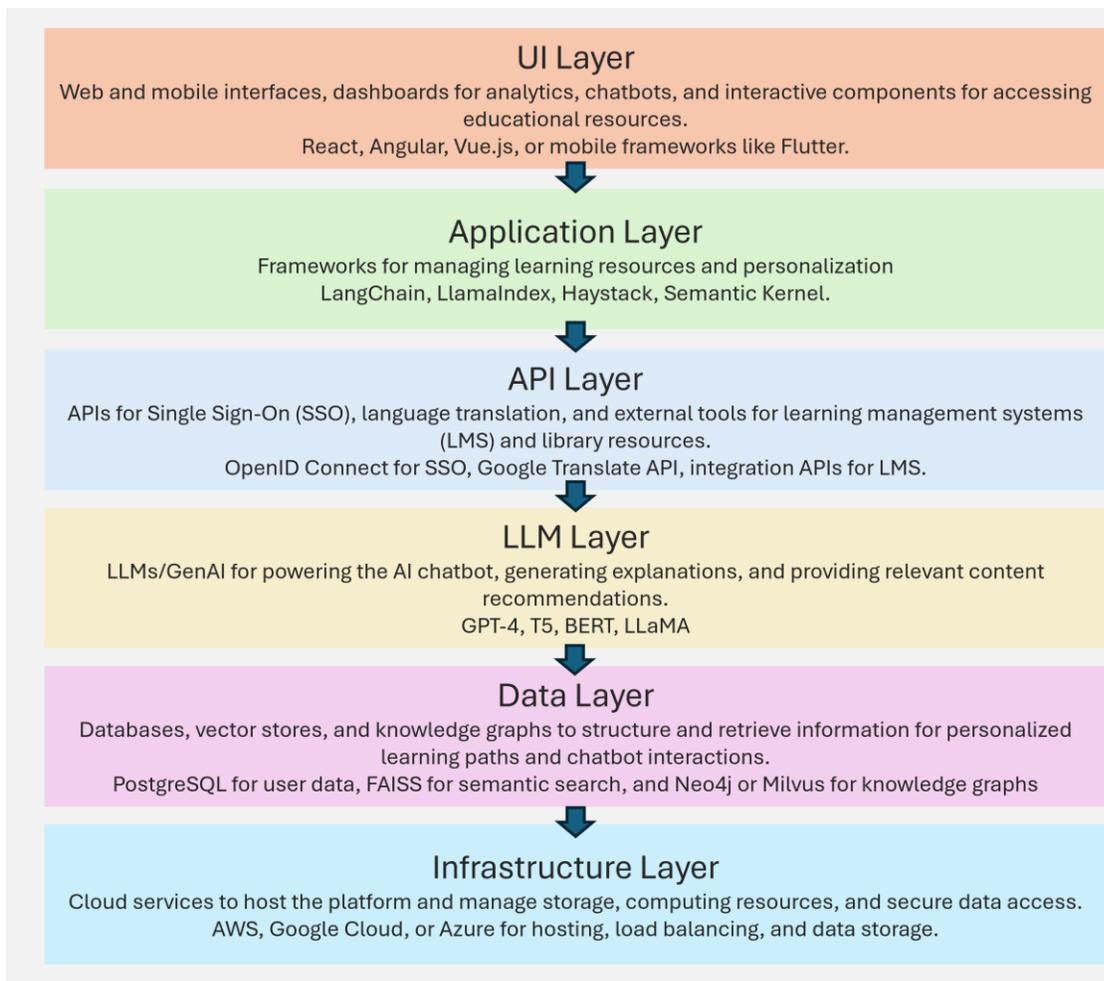

*Fig. 3 Proposed GenAI Technology Stack for Educational Platforms*

At the heart of this ecosystem (Fig. 2) is an AI-powered chatbot, enhanced with a Retrieval-Augmented Generation (RAG) capability. This component serves as an intelligent, context-aware assistant that provides tailored responses by combining domain-specific knowledge from institutional repositories with the generative power of LLMs (e.g., Mistral 7B).

In this setup, the RAG-based chatbot connects with structured resources (e.g., online library access, LMS, analytics dashboards, publication platforms) through secure cloud services. It leverages preprocessed data stored in a Domain Data Lake, which is converted into vector embeddings using an embedding model (like OpenAI's). These are stored in a Vector Database, enabling semantic retrieval. When a student or staff member queries the chatbot, it pulls relevant content, augments the prompt, and passes it to the LLM for response generation—supporting tasks like personalized study help, automated literature reviews, or administrative Q&A.

In more detail, Fig. 4 showcases a model information retrieval system integrating RAG with LLMs to provide responses to user queries from a chat interface. This diagram illustrates how various components from data preprocessing to advanced natural language processing and LLM integration work together to provide enhanced information retrieval and interaction. The following description for each module of this framework:

- **Data management and preprocessing**: Data management and preprocessing module is responsible for handling various data preprocessing tasks such as data cleaning, crawling, and gathering data from diverse data sources. Data is stored in a Domain Data Lake, which consists of structured and unstructured data (PDFs, HTML, etc.). The legal and ethical guidelines, represented by Business Rules, inform the pipeline to ensure compliance.

- **Embedding Model**: Text from the data lake is divided into smaller chunks, which are then transformed into numerical representations, or embeddings, using an embedding model. Embedding model utilizes OpenAI embeddings API to generate embeddings from this preprocessed chunked text. The embedding vector is calculated for each chunk , which are then stored in the Vector DB for fast retrieval.

- **Retrieval from Vector DB**: The embeddings are saved in a Vector DB for the text chunks along with their associated content. It facilitates fast search and retrieval based on vector similarity. When a query is initiated, the system retrieves relevant documents that match the embedding vector.

- **Augment and Generate**: The chat UI forwards the user query to the system and relevant information retrieved from the Vector DB (based on embeddings) is used to augment the query before passing it to the LLM for further processing. This process can be termed as Prompt augmentation. The query and retrieved information are then sent to the LLM (Mistral 7B in this case). The LLM generates a refined response, combining both the user's input and the augmented information from the Vector DB.

- **LLM Fine-tuning**: The system optionally supports fine-tuning the LLM with domain-specific data for improved accuracy. Fine-tuning involves additional tasks like formatting, testing, and preprocessing data (e.g., data cleaning, ensuring the model is trained on GPUs for computational efficiency).

- **User Interaction and Flow using Chat UI**: The user interacts with the system through a conversational interface and then the system processes queries, augments them with retrieved documents, and delivers AI-generated responses back to the user.

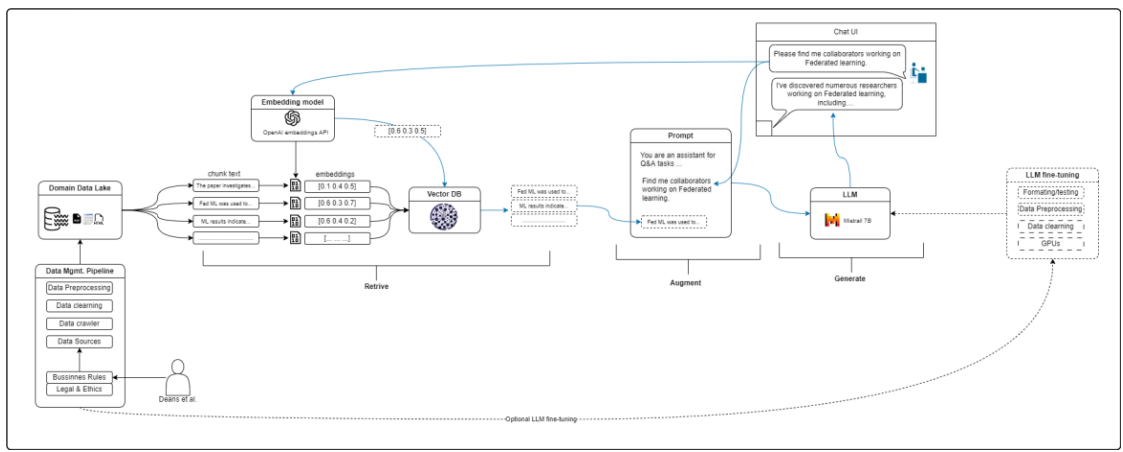

*Fig. 4 Chatbot Representation*

## 5.2 Governance, Risk, and Compliance in GenAI Education Platforms

In light of the EU AI Act, the deployment of GenAI systems in educational settings necessitates careful consideration of regulatory compliance, ethical governance, and institutional accountability. The proposed AI-powered educational platform—featuring adaptive learning, RAG-enabled chatbots, automated assessments, and AI-supported publication tools—likely qualifies as a high-risk AI system, as defined under **Title III, Chapter 1, Article 6** of the EU AI Act. High-risk systems are those that may significantly affect the rights, safety, or well-being of individuals, especially when used in educational or vocational training contexts where they influence access to education or career outcomes (Annex III, Section 3 (European Commission, 2024a)).

To ensure full compliance, the platform must address the following key requirements mandated by the EU AI Act:

**1. Risk Classification and Compliance Obligations (*Articles 6 (European Union, 2024g) and 8–15*)**
Since the platform provides automated learning assessments, student profiling, and decision-support tools in educational settings, it falls under the Annex III list of high-risk applications. This classification triggers mandatory requirements including:

- **Risk Management System** (*Article 9 (European Union, 2024h)*): A continuous, documented risk management process must be established to identify, evaluate, and mitigate risks associated with the AI system during its lifecycle. This includes model performance monitoring, edge case testing, and fallback mechanisms in case of AI failure or inaccuracy.

- **Data and Data Governance** (*Article 10 (European Union, 2024a)*): Training and evaluation datasets used for LLMs and recommender systems must be relevant, representative, and free from bias to ensure fairness and accuracy. Preprocessing pipelines must include bias detection, data cleansing, and documentation of dataset provenance.

- **Technical Documentation** (*Article 11 (European Union, 2024b)*): Detailed documentation must be maintained, including system architecture, intended purpose, design specifications, and validation metrics to support transparency and future audits by competent authorities.

- **Record-Keeping (Logs)** (*Article 12 (European Union, 2024c)*): Systems must log operations to allow traceability and auditing. This is especially important for adaptive assessments and personalized learning paths that affect grading or educational outcomes.

- **Transparency and Provision of Information to Users** (*Article 13 (European Union, 2024d)*): End users—including students and staff—must be explicitly informed that they are interacting with an AI system, and must receive clear instructions on how to interpret the output (e.g., feedback, recommendations) and how to seek human assistance.

- **Human Oversight** (*Article 14 (European Union, 2024e)*): The platform must ensure human oversight mechanisms are in place to allow intervention, override of AI decisions, and avoidance of overreliance. For example, educators must be able to review and amend AI-generated feedback or grades.

- **Accuracy, Robustness, and Cybersecurity** (*Article 15 (European Union, 2024f)*): The system must demonstrate a high level of accuracy, resilience against manipulation, and protection from adversarial inputs. This includes ensuring the chatbot and analytics dashboard cannot be misled by ambiguous queries or injected content.

**2. GDPR and Ethical Use of AI in Education**

In addition to the AI Act, the platform must comply with the GDPR (European Union, 2016), especially concerning:

- **Data Minimization and Purpose Limitation:** Only data necessary for specific learning or administrative functions should be collected and processed (*GDPR Articles 5–6 (GDPR, 2016h, 2016i)*).

- **User Consent and Rights:** Students must be informed about data usage and retain rights to access, correct, or delete personal information (*GDPR Articles 7 (GDPR, 2016j), 12–18 (GDPR, 2016a, 2016b, 2016c, 2016d, 2016e, 2016f)*).

- **Privacy-by-Design:** The system architecture must incorporate privacy-enhancing technologies and restrict access based on defined roles via secure identity management (e.g., through Keycloak RBAC and SSO) (*GDPR Article 25 (GDPR, 2016g)*).

### 3. Intellectual Property, Plagiarism, and AI-Generated Content Transparency

The platform's AI-assisted publication tools and LLM-enabled writing support raise unique challenges around authorship, academic integrity, and intellectual property:

- **AI-Generated Content Disclosure:** When generative AI contributes to content creation—such as abstracts, summaries, or citations—there must be clear attribution and disclaimers, in accordance with *Article 52* of the EU AI Act (EU AI Act, 2025), which mandates transparency for general-purpose AI systems and content labeling to avoid user deception. While the Act does not explicitly require detailed labeling of every AI-generated content piece, it emphasizes that users should be made aware that content or interactions involve AI to maintain trust and uphold transparency standards (European Commission, 2021)

- **Plagiarism Detection and Source Attribution:** Integrating AI-based plagiarism detection and proper citation management supports compliance with scholarly standards and helps prevent misuse of generative tools. Intellectual property boundaries must also be respected, particularly when AI uses or replicates copyrighted content without proper licensing.

### 4. AI System Registration and CE Marking (*Articles 16–23*)

Before deployment in EU-based institutions, the platform may be subject to:

- **Conformity Assessment Procedures:** As a high-risk system, it must undergo internal or third-party conformity checks to demonstrate compliance (*Articles 19–21*).

- **Registration in the EU AI Database:** High-risk systems must be registered in the EU's publicly accessible AI database, including detailed information about their purpose, capabilities, and risk mitigation strategies (*Article 51*).

- **CE Marking:** Upon compliance, the system must bear the CE marking to affirm conformity with EU AI Act requirements (*Article 49*).

### 5. Institutional Strategy and Trust-Building

Finally, for sustainable and ethical adoption, institutions should establish a broader AI governance strategy that includes:

- **Ethics Committees or AI Advisory Boards** to evaluate AI deployments, in line with the risk-based approach recommended by the EU AI Act (*Recital 60* (European Commission, 2024b)).

- **User Training and Digital Literacy Programs** to prepare staff and students for interacting with AI-driven tools responsibly.

- **Feedback and Redress Mechanisms** allowing users to challenge AI decisions and seek clarification or correction, aligned with transparency and accountability principles (*Articles 13, 14*).

In conclusion, integrating GenAI into education offers significant potential, but it must be executed within a strong framework of legal, ethical, and institutional safeguards. Adhering to the EU AI Act not only ensures compliance but also builds user trust and future-proofs the platform against evolving regulatory landscapes.

# 6 Discussion

This section synthesizes key findings from the staff and student surveys, which explored stakeholders' perceptions and ideas regarding the integration of GenAI in higher education. Building on these findings, we distil a set of high-level requirements and introduce a conceptual framework designed to support responsible and effective GenAI integration. We conclude by acknowledging the study's limitations and outlining directions for future research.

## 6.1 Stakeholder Perceptions of GenAI

A majority of staff and student respondents anticipated that GenAI chatbots would enable personalized learning features such as 24/7 learning support and personalized learning paths. Both groups thought GenAI chatbots could be used for searching papers or books.

While both groups shared some similar visions and preferences, we also notice differences between the groups, highlighting the importance of addressing different needs when integrating GenAI in higher education. For instance, a majority of staff respondents thought chatbots could be used for answering general questions about courses, whereas only about 1/5 of student respondents preferred this type of assistance. This may be due to students' concerns about the quality of GenAI responses and the need for student-teacher interaction.

The surveys revealed that more staff respondents than student respondents used GenAI daily. Additionally, the staff used GenAI in more tasks than the students. It could be due to students' concerns about violating institutional policies. Potential differences between existing policies might also limit students' use of GenAI, e.g., some courses require reporting on the use of GenAI, while other courses might prohibit its use at all. It is also possible that some student respondents might prefer to disclose only part of their GenAI usage, due to concerns about staff expectations. These possible reasons further highlight the importance of trust, clear institutional policies, and effective communication among the stakeholders, in order to harness the technology optimally.

While staff and student respondents reported productivity boosts from using GenAI, they were also concerned about its limitations and potential negative impacts. In fact, we found concerns in open-ended responses from nearly every section.

Among identified concerns, most respondents from both groups were very concerned about the quality of GenAI responses. Additionally, staff were also very concerned about privacy and data ownership, academic integrity, and copyright infringement. Staff had voiced their specific concern about the negative impacts of students' use of GenAI in writing, expressing concern that it may hinder learning. On the other hand, student respondents were more concerned about GenAI's impact on their creativity and becoming overly dependent on the technology.

The dual perceptions on the benefits and concerns of GenAI emerged from our surveys are consistent with other survey studies in the recent years (Alharbi, 2024; Chan & Hu, 2023; Jin et al., 2025; Watermeyer et al., 2023; Yusuf et al., 2024), underscoring the ongoing challenge of harnessing the technology while addressing concerns. Despite the challenge, many have anticipated that GenAI can transform higher education. However, the responsible integration of GenAI into higher education calls for stakeholder engagement, clear policies, requirements, mitigative strategies, and continuous monitoring.

## 6.2 Identified High-Level Requirements

The surveys surfaced several high-level requirements that can inform the effective integration of GenAI into higher education. In particular, asking respondents to propose ideas for GenAI applications enabled the identification of specific, potentially high-impact functional requirements relevant to them. This approach recognizes that requirements can be influenced by cultural, disciplinary, or institutional contexts, aligning with Yusuf et al. (2024), who stress the importance of considering cultural aspects when designing GenAI strategies and policies. Below, we summarize the main functional and non-functional requirements that emerged from the surveys.

### 6.2.1 Functional Requirements

For learning, the majority of both staff and student respondents perceived value in using GenAI for information search and extraction, indicating their awareness of its capabilities. In addition, most student respondents considered programming assistance to be quite or very useful for enhancing their coding skills, representing a disciplinary-specific requirement in the field of Information Technology.

For teaching, several staff respondents highlighted the potential of GenAI to create engaging teaching content and streamline course assessments. These requirements align with the expectations that GenAI can enhance student engagement and accommodate diverse learning needs and styles (Kasneci et al., 2023). In addition, early evidence encourages further exploration of GenAI's potential to transform education (Deng et al., 2025). However, while most staff respondents viewed GenAI chatbots as useful for answering course-related questions, student respondents were clearly less enthusiastic. Without identifying and addressing students' concerns (e.g., accuracy and the need for human interaction), the adoption of GenAI chatbots for this purpose risks resistance and ineffective resource allocation.

For research and development, most staff respondents agreed that using GenAI for literature reviews and software prototyping would be quite or very useful. Regarding literature reviews, Issa et al. (2024) reported that earlier models such as ChatGPT-3.5 performed poorly in classifying paragraphs but were able to provide helpful explanations about the reviewed content. They also recommended exploring the capabilities of newer models for such tasks. Commercial tools like Scopus AI, which support research paper search and summarization, further demonstrate the growing potential in this area. For programming tasks, tools such as GitHub Copilot are already in use. These examples suggest the feasibility of GenAI in streamlining both literature reviews and software development. Collectively, these findings call for further requirement analysis and evaluation of existing tools to inform institutional adoption. Beyond these common themes, individual respondents also suggested using GenAI to support research design and data analysis.

For administration, streamlining research and funding application processes, as well as supporting international student admissions, were identified from staff responses as potentially high-value requirements. While previous survey studies have noted that staff perceive GenAI as useful for administrative or routine tasks (Cervantes et al., 2024; Lee et al., 2024; Watermeyer et al., 2023), the identified requirements in this study provide more concrete examples of how GenAI could enhance administrative workflows in higher education.

## 6.2.2 Non-Functional Requirements

In order to realize and optimize the value of GenAI applications in higher education, such as those presented in the functional requirements, we agree with Al-Omari et al. (2025) in emphasizing that ethical and pedagogical concerns must also be identified and addressed. While AI literacy, institutional policies, and governance are essential (Chan, 2023), non-functional requirements translate and operationalize stakeholders' expectations regarding how such systems should function in practice. Below, we present key non-functional requirements that define the conditions of responsible and effective GenAI integration in higher education. These requirements are rooted in the concerns and expectations expressed by survey respondents and supported by existing literature. As with the functional requirements, their relevance may vary depending on institutional, disciplinary, or cultural contexts.

*GenAI applications in higher education must deliver response quality appropriate to their intended use.* Their unique capability of generating new content based on prompts is both a strength and a limitation. While this enables flexible and meaningful interaction, it also raises concerns: these models do not "know" whether their outputs are accurate, as responses are based on predictions shaped by model architecture, training data, and prompts (Mittal et al., 2024). The majority of both staff and student respondents expressed strong concern about the quality of GenAI responses, including inaccurate, biased, or inappropriate outputs. These concerns align with ongoing discussions in the literature (Chan & Hu, 2023; Issa et al., 2024; Jin et al., 2025; Kasneci et al., 2023). Although no model guarantees perfect accuracy, early evidence suggests GenAI can still enhance student learning when used thoughtfully (Deng et al., 2025). Rather than expecting perfect accuracy, we suggest that institutions articulate clear, context-specific criteria for acceptable response quality. For example, requiring citations and links in tools used in information search.

*These applications should be designed to support and protect academic integrity by addressing risks of misuse*. Many staff and student respondents expressed serious concern that GenAI could compromise academic integrity. Staff specifically articulated their worries about the erosion of research integrity, as well as the potential for student cheating. These concerns are echoed by researchers, educators, and students globally (Alharbi, 2024; Kasneci et al., 2023; Lee et al., 2024; Watermeyer et al., 2023; Yusuf et al., 2024). In teaching, staff respondents noted that they are already updating assessment methods in response to GenAI. GenAI applications for student use should be designed in alignment with these efforts. In research and administration, the responsibility for upholding academic integrity rests primarily with staff. Accordingly, GenAI tools should be guided by institutional policies and intentionally designed to uphold ethical standards across academic contexts.

*These applications, when intended to support more advanced learning or complex tasks, should be designed to promote higher-order thinking*. This encompasses more sophisticated cognitive processes such as analysis, evaluation, and creation (Krathwohl, 2002). Many staff and student respondents were very concerned about the potential negative impacts of GenAI on learning, critical thinking, and communication skills. Additionally, students were particularly worried about diminishing creativity and becoming overly dependent on GenAI. These concerns challenge higher education to continuously investigate and innovate effective ways of integrating GenAI (Chan & Hu, 2023; Jin et al., 2025; Lee et al., 2024). Meanwhile, empirical studies have highlighted the potential of using GenAI to foster higher-order thinking and communication skills when properly designed. For example, instead of delegating writing tasks to ChatGPT, Garcia-Varela et al. (2025) showed that it could be prompted to train teaching staff to write more effective feedback for students. Similarly, Zhang et al. (2025) reported the feasibility of engaging students in developing logical thinking and improving argumentative writing with GenAI assistance. In computer science, Kohen-Vacs et al. (2025) asked students to review GenAI generated code, fostering problem-solving and critical thinking skills. In research contexts, Borge et al. (2024) showed that GenAI can be instructed to help researchers think critically about their work. These examples illustrate promising avenues for using GenAI to support, rather than diminish higher-order thinking.

*These applications, regardless of their specific use cases, must protect users' privacy and data properly*. Privacy and data ownership were among the top concerns raised by both staff and student respondents. These concerns are echoed by other educators and students (Bernabei et al., 2023; Cervantes et al., 2024; Chan & Hu, 2023). We argue that privacy and data protection should be considered a mandatory non-functional requirement to ensure compliance with institutional policies as well as relevant regulations such as GDPR. Many reported educational applications use ChatGPT directly (Deng et al., 2025), which may not meet stricter data protection requirements in sensitive contexts such as medical research. However, with additional effort, enhancing privacy and data protection is achievable. For example, Ng et al. (2025) demonstrated the implementation of a private GenAI environment that allowed medical researchers to explore applications in fall detection and psychological assessment while safeguarding their data.

*These applications should facilitate personalized interaction experiences through system prompt and other forms of context management.* Consistent with previous studies (Chan & Hu, 2023; Deng et al., 2025; Yusuf et al., 2024), staff and student respondents anticipated that GenAI could offer on-demand support and enhance personalized learning experiences. Most respondents reported using GenAI tools to answer their questions, such as explaining concepts and terminologies. Particularly in the context of Information Technology, individual programming assistance emerged as a popular use case. However, when asked about using GenAI for other specific personal tasks, such as making a personal study plan or providing feedback on assignments or projects, respondents expressed less preference and perceived usefulness. These subtle differences between general anticipation of personalized interactions and hesitations around specific personal tasks may reflect underlying concerns regarding unreliable response quality, the risk to academic integrity, diminishing higher-order thinking, and especially, privacy. Han et al. (2025) highlight the importance of supporting student autonomy and data control. We further argue that personalized interaction experiences can be enabled without collecting or storing unnecessary personal information upfront. For example, when combined with adequate user privacy and data protection measures, personalization could be facilitated by allowing users to specify the focus and style of their interaction and manage their context (e.g., including or removing their documents and previous chat messages).

*These applications must support institutional governance and comply with relevant regulations by enabling auditing and accountability.* While staff and student respondents recognized the benefits of GenAI, they also raised strong concerns, including the quality of GenAI responses, risks to academic integrity, diminishing higher-order thinking, privacy and data protection, security, and copyright infringements, highlighting the critical need for responsible integration of GenAI. Similarly, researchers have drawn attention to the need for effective institutional governance (Chan, 2023; Chan & Hu, 2023; Yusuf et al., 2024) and regulatory compliance (Al-Omari et al., 2025; Ng et al., 2025). Building on these insights, we articulate this non-functional requirement as a necessary consideration when adopting or developing GenAI applications for higher education. For example, GenAI applications should provide stakeholders with key performance indicators in a privacy-preserving manner, such as through anonymous feedback mechanisms. This information can support institutional governance by enabling stakeholders to monitor system performance, ensure regulatory compliance, and assess impacts, as a result facilitating informed decision-making regarding the responsible and effective use of GenAI.

To conclude, in this section we presented potentially high-value functional requirements for learning, teaching, research, and administration, highlighting the potential of GenAI to transform Information Technology and Electrical Engineering education. We also presented key non-functional requirements addressing the major concerns raised by respondents. While some requirements reflect the specific cultural, institutional, and disciplinary context of the survey, such as leveraging GenAI for programming support, many address broader needs and concerns in higher education, including engaging students in learning and upholding academic integrity. Building on these requirements, the next section discusses our framework that can help operationalize responsible GenAI integration at an institutional level.

## 6.3 Conceptual Framework

Integrating GenAI technologies into higher education platforms presents transformative opportunities to enhance personalized learning, streamline administrative workflows, and empower educators in their teaching and research endeavors. The proposed GenAI-powered ecosystem anchored by adaptive learning paths, intelligent chatbots, and comprehensive analytics demonstrates a practical and scalable approach to bridging diverse stakeholder needs with cutting-edge AI capabilities.

However, as highlighted by the EU AI Act and GDPR regulatory frameworks, the deployment of such high-risk AI systems demands rigorous governance, ethical oversight, and transparency to safeguard the rights, privacy, and well-being of students, educators, and administrators. Compliance with risk management protocols, data governance principles, and human oversight mechanisms is not merely a legal obligation but a foundational element for fostering trust and acceptance among users.

Moreover, the challenges of intellectual property, AI-generated content transparency, and academic integrity necessitate the integration of robust plagiarism detection and clear attribution practices. Ensuring that users understand the involvement of AI in content creation and decision-making processes is critical for maintaining institutional credibility and upholding scholarly standards.

Beyond regulatory compliance, institutions must embed AI literacy, user empowerment, and feedback channels within their AI governance frameworks. This holistic strategy, including ethics committees and continuous training programs, cultivates an environment where AI augments human expertise rather than replacing it, allowing educators and students to harness AI tools effectively and responsibly.

In essence, the successful adoption of GenAI in education hinges on balancing innovation with accountability, personalization with fairness, and automation with human oversight. By embedding these principles at the core of design and deployment, educational institutions can unlock the full potential of AI technologies while ensuring ethical, transparent, and sustainable outcomes that align with evolving societal and regulatory expectations.

This multi-faceted approach not only future-proofs educational platforms against regulatory shifts but also positions them as trusted enablers of knowledge, equity, and academic excellence in the digital age.

## 6.4 Limitations and Future Research

When interpreting the study findings, the following limitations should be considered. Firstly, due to the small sample size, some relevant concerns may remain unidentified or underrepresented. Secondly, while many non-functional requirements identified in this study could be relevant to other higher education contexts, some functional requirements, such as programming assistance, are specific to the field of Information Technology. Thirdly, the proposed framework is primarily informed by survey findings and literature and requires further validation through implementation and evaluation. These limitations necessitate the consideration of the specific context of the study when interpreting the results.

Building on what we have learned from this study, we identify the following avenues for future research. Firstly, more empirical evidence is needed to evaluate the impacts of GenAI across the core functions of higher education to better inform institutional decision-making. For example, future studies could apply and elaborate on the identified requirements and framework, gathering further evidence through real-world implementation. Secondly, resonating with ongoing discourse, this study identified critical stakeholder concerns such as accuracy, academic integrity, higher-order thinking, and privacy. These concerns call for future research in light of rapid technological evolution, with the goal of developing strategies and solutions that enable higher education institutions to harness GenAI effectively and responsibly.

# 7 Conclusion

The rapid development and disruptive impact of GenAI are reshaping higher education in unprecedented ways. While staff and students have begun adopting tools such as ChatGPT, perceptions remain mixed, and ethical and pedagogical concerns persist. These perceptions can also be influenced by cultural factors (Yusuf et al., 2024). Moreover, supranational regulations like the EU AI Act require institutions to ensure compliance when integrating GenAI and other cognitive systems. These factors highlight the importance of engaging stakeholders and aligning AI integration with regulatory obligations.

To address this need, we surveyed staff and students from the Faculty of Information Technology and Electrical Engineering to explore their concerns and expectations, identify high-level requirements, and propose a conceptual framework for integration that accounts for both stakeholder needs and regulatory compliance.

The survey findings show that while many respondents recognized the productivity benefits of GenAI, significant concerns remain, particularly regarding GenAI response quality, data privacy, impacts on higher-order thinking, and risks to academic integrity. Additionally, we identified a set of functional and non-functional requirements that reflect both discipline-specific needs, such as programming support, and broader concerns discussed in previous studies, including the protection of academic integrity, promotion of higher-order thinking, data protection, and regulatory alignment. Conceptually, the study contributes to the ongoing discussion about responsible AI integration in higher education by illustrating how stakeholder-informed requirements and regulation compliance must inform institutional integration and governance. Building on these insights, we developed a conceptual framework that explicitly considers the requirements of the EU AI Act and GDPR. Together, these contributions offer both theoretical grounding and practical guidance for institutions seeking to design and implement GenAI tools responsibly.

Given the limited sample size and the specific disciplinary scope of this study, the findings should be interpreted in context and not generalized across all academic environments. Future research should further investigate ethical and pedagogical implications of GenAI, develop effective integration strategies, and evaluate them across diverse higher education contexts.

# References

Airaj, M. (2024). Ethical artificial intelligence for teaching-learning in higher education. *Education and Information Technologies*, *29*(13), 17145–17167. https://doi.org/10.1007/s10639-024-12545-x

Alemdag, E. (2023). The effect of chatbots on learning: A meta-analysis of empirical research. *Journal of Research on Technology in Education*, *0*(0), 1–23. https://doi.org/10.1080/15391523.2023.2255698

Aler Tubella, A., Mora-Cantallops, M., & Nieves, J. C. (2023). How to teach responsible AI in Higher Education: Challenges and opportunities. *Ethics and Information Technology*, *26*(1), 3. https://doi.org/10.1007/s10676-023-09733-7

Alharbi, W. (2024). Mind the Gap, Please!: Addressing the Mismatch Between Teacher Awareness and Student AI Adoption in Higher Education. *International Journal of Computer-Assisted Language Learning and Teaching (IJCALLT)*, *14*(1), 1–28. https://doi.org/10.4018/IJCALLT.351245

Aljohani, N. R., Daud, A., Abbasi, R. A., Alowibdi, J. S., Basheri, M., & Aslam, M. A. (2019). An integrated framework for course adapted student learning analytics dashboard. *Computers in Human Behavior*, *92*, 679–690. https://doi.org/10.1016/j.chb.2018.03.035

Al-Omari, O., Alyousef, A., Fati, S., Shannaq, F., & Omari, A. (2025). Governance and Ethical Frameworks for AI Integration in Higher Education: Enhancing Personalized Learning and Legal Compliance. *Journal of Ecohumanism*, *4*(2), 80–86. https://doi.org/10.62754/joe.v4i2.5781

An, S., Zhang, S., Guo, T., Lu, S., Zhang, W., & Cai, Z. (2025). Impacts of generative AI on student teachers' task performance and collaborative knowledge construction process in mind mapping-based collaborative environment. *Computers & Education*, *227*, 105227. https://doi.org/10.1016/j.compedu.2024.105227

Ansari, A. N., Ahmad, S., & Bhutta, S. M. (2023). Mapping the global evidence around the use of ChatGPT in higher education: A systematic scoping review. *Education and Information Technologies*. https://doi.org/10.1007/s10639-023-12223-4

Bahassi, H., Azmi, M., & Khiat, A. (2024). Cognitive systems for education: Architectures, innovations, and comparative analyses. *Procedia Computer Science*, *238*, 436–443.

Balan, A. (2024). Examining the ethical and sustainability challenges of legal education's AI revolution. *International Journal of the Legal Profession*, *31*(3), 323–348. https://doi.org/10.1080/09695958.2024.2421179

Balasubramanian, P., Liyana, S., Sankaran, H., Sivaramakrishnan, S., Pusuluri, S., Pirttikangas, S., Peltonen, E., (2025). Generative ai for cyber threat intelligence: applications, challenges, and analysis of real-world case studies. *Artificial Intelligence Review* 58, 336.

Bengesi, S., El-Sayed, H., Sarker, M. K., Houkpati, Y., Irungu, J., & Oladunni, T. (2024). Advancements in Generative AI: A Comprehensive Review of GANs, GPT, Autoencoders, Diffusion Model, and Transformers. *IEEE Access*, *12*, 69812–69837. https://doi.org/10.1109/ACCESS.2024.3397775

Bernabei, M., Colabianchi, S., Falegnami, A., & Costantino, F. (2023). Students' use of large language models in engineering education: A case study on technology acceptance, perceptions, efficacy, and detection chances. *Computers and Education: Artificial Intelligence*, *5*, 100172. https://doi.org/10.1016/j.caeai.2023.100172

Borge, M., Smith, B. K., & Aldemir, T. (2024). Using generative ai as a simulation to support higher-order thinking. *International Journal of Computer-Supported Collaborative Learning*, *19*(4), 479–532. https://doi.org/10.1007/s11412-024-09437-0

Burk-Rafel, J., & Triola, M. M. (2025). Precision Medical Education: Institutional Strategies for Successful Implementation. *Academic Medicine*. https://doi.org/10.1097/ACM.0000000000005980

Capuano, N., & Toti, D. (2019). Experimentation of a smart learning system for law based on knowledge discovery and cognitive computing. *Computers in Human Behavior*, *92*, 459–467. https://doi.org/10.1016/j.chb.2018.03.034

Cervantes, J., Smith, B., Ramadoss, T., D'Amario, V., Shoja, M. M., & Rajput, V. (2024). Decoding medical educators' perceptions on generative artificial intelligence in medical education. *Journal of Investigative Medicine*, *72*(7), 633–639. https://doi.org/10.1177/10815589241257215

Chan, C. K. Y. (2023). A comprehensive AI policy education framework for university teaching and learning. *International Journal of Educational Technology in Higher Education*, *20*(1), 38. https://doi.org/10.1186/s41239-023-00408-3

Chan, C. K. Y., & Hu, W. (2023). Students' voices on generative AI: Perceptions, benefits, and challenges in higher education. *International Journal of Educational Technology in Higher Education*, *20*(1), 43. https://doi.org/10.1186/s41239-023-00411-8

Coniam, D. (2008). Evaluating the language resources of chatbots for their potential in English as a second language. *ReCALL*, *20*(1), 98–116. https://doi.org/10.1017/S0958344008000815

Dabis, A., & Csáki, C. (2024). AI and ethics: Investigating the first policy responses of higher education institutions to the challenge of generative AI. *Humanities and Social Sciences Communications*, *11*(1), 1006. https://doi.org/10.1057/s41599-024-03526-z

Deng, R., Jiang, M., Yu, X., Lu, Y., & Liu, S. (2025). Does ChatGPT enhance student learning? A systematic review and meta-analysis of experimental studies. *Computers & Education*, *227*, 105224. https://doi.org/10.1016/j.compedu.2024.105224

EU AI Act. (2025). *Key issue 5: Transparency obligations*. https://www.euaiact.com/key-issue/5

European Commission. (2021). *Proposal for a regulation laying down harmonised rules on artificial intelligence (artificial intelligence act)*. EUR-Lex. https://eur-lex.europa.eu/legal-content/EN/TXT/?uri=CELEX%3A52021PC0206#document2

European Commission, (2024a). *Annex III – high-risk AI systems*. https://eur-lex.europa.eu/legal-content/EN/TXT/?uri=CELEX:52021PC0206#d1e32-67-1

European Commission, (2024b). *Recital 60 – governance and oversight in high-risk AI systems*. https://eur-lex.europa.eu/legal-content/EN/TXT/?uri=CELEX:52021PC0206#d1e32-136-1

European Commission. (2025). *Regulatory framework on artificial intelligence (AI)*. https://digital-strategy.ec.europa.eu/en/policies/regulatory-framework-ai

European Union. (2016). *General data protection regulation (GDPR) – regulation (EU) 2016/679*. https://eur-lex.europa.eu/eli/reg/2016/679/oj

European Union. (2024a). *Artificial intelligence act - article 10: Data and data governance*. https://artificialintelligenceact.eu/article/10/

European Union. (2024b). *Artificial intelligence act - article 11: Technical documentation*. https://artificialintelligenceact.eu/article/11/

European Union. (2024c). *Artificial intelligence act - article 12: Record-keeping*. https://artificialintelligenceact.eu/article/12/

European Union. (2024d). *Artificial intelligence act - article 13: Transparency and provision of information to deployers*. https://artificialintelligenceact.eu/article/13/

European Union. (2024e). *Artificial intelligence act - article 14: Human oversight*. https://artificialintelligenceact.eu/article/14/

European Union. (2024f). *Artificial intelligence act - article 15: Accuracy, robustness and cybersecurity*. https://artificialintelligenceact.eu/article/15/

European Union. (2024g). *Artificial intelligence act - article 6*. https://artificialintelligenceact.eu/article/6/

European Union. (2024h). *Artificial intelligence act - article 9: Risk management system*. https://artificialintelligenceact.eu/article/9/

Garcia-Varela, F., Bekerman, Z., Nussbaum, M., Mendoza, M., & Montero, J. (2025). Reducing interpretative ambiguity in an educational environment with ChatGPT. *Computers & Education*, *225*, 105182. https://doi.org/10.1016/j.compedu.2024.105182

GDPR. (2016a). *General data protection regulation (GDPR) – article 12: Transparent information, communication and modalities for the exercise of the rights of the data subject*. https://gdpr-info.eu/art-12-gdpr/

GDPR. (2016b). *General data protection regulation (GDPR) – article 13: Information to be provided where personal data are collected from the data subject*. https://gdpr-info.eu/art-13-gdpr/

GDPR. (2016c). *General data protection regulation (GDPR) – article 14: Information to be provided where personal data have not been obtained from the data subject*. https://gdpr-info.eu/art-14-gdpr/

GDPR. (2016d). *General data protection regulation (GDPR) – article 15: Right of access by the data subject*. https://gdpr-info.eu/art-15-gdpr/

GDPR. (2016e). *General data protection regulation (GDPR) – article 16: Right to rectification*. https://gdpr-info.eu/art-16-gdpr/

GDPR. (2016f). *General data protection regulation (GDPR) – article 17: Right to erasure ('right to be forgotten')*. https://gdpr-info.eu/art-17-gdpr/

GDPR. (2016g). *General data protection regulation (GDPR) – article 25: Data protection by design and by default*. https://gdpr-info.eu/art-25-gdpr/

GDPR. (2016h). *General data protection regulation (GDPR) – article 5: Principles relating to processing of personal data*. https://gdpr-info.eu/art-5-gdpr/

GDPR. (2016i). *General data protection regulation (GDPR) – article 6: Lawfulness of processing*. https://gdpr-info.eu/art-6-gdpr/

GDPR. (2016j). *General data protection regulation (GDPR) – article 7: Conditions for consent*. https://gdpr-info.eu/art-7-gdpr/

Guan, L., Zhang, E. Y., & Gu, M. M. (2024). Examining generative AI–mediated informal digital learning of English practices with social cognitive theory: A mixed-methods study. *ReCALL*, 1–17. https://doi.org/10.1017/S0958344024000259

Guo, K., Pan, M., Li, Y., & Lai, C. (2024). Effects of an AI-supported approach to peer feedback on university EFL students' feedback quality and writing ability. *The Internet and Higher Education*, *63*, 100962. https://doi.org/10.1016/j.iheduc.2024.100962

Han, B., Nawaz, S., Buchanan, G., & McKay, D. (2025). Students' Perceptions: Exploring the Interplay of Ethical and Pedagogical Impacts for Adopting AI in Higher Education. *International Journal of Artificial Intelligence in Education*. https://doi.org/10.1007/s40593-024-00456-4

Hashmi, N., & Bal, A. S. (2024). Generative AI in higher education and beyond. *Business Horizons*.

Hu, W., Tian, J., & Li, Y. (2025). Enhancing student engagement in online collaborative writing through a generative AI-based conversational agent. *The Internet and Higher Education*, *65*, 100979. https://doi.org/10.1016/j.iheduc.2024.100979

Hughes, L., Malik, T., Dettmer, S., Al-Busaidi, A. S., & Dwivedi, Y. K. (2025). Reimagining Higher Education: Navigating the Challenges of Generative AI Adoption. *Information Systems Frontiers*. https://doi.org/10.1007/s10796-025-10582-6

Issa, M., Faraj, M., & AbiGhannam, N. (2024). Exploring ChatGPT's Ability to Classify the Structure of Literature Reviews in Engineering Research Articles. *IEEE Transactions on Learning Technologies*, *17*, 1819–1828. https://doi.org/10.1109/TLT.2024.3409514

Jeon, J., Lee, S., & Choe, H. (2023). Beyond ChatGPT: A conceptual framework and systematic review of speech-recognition chatbots for language learning. *Computers & Education*, *206*, 104898. https://doi.org/10.1016/j.compedu.2023.104898

Jin, F., Sun, L., Pan, Y., & Lin, C.-H. (2025). High Heels, Compass, Spider-Man, or Drug? Metaphor Analysis of Generative Artificial Intelligence in Academic Writing. *Computers & Education*, 105248. https://doi.org/10.1016/j.compedu.2025.105248

Kasneci, E., Sessler, K., Küchemann, S., Bannert, M., Dementieva, D., Fischer, F., Gasser, U., Groh, G., Günnemann, S., Hüllermeier, E., Krusche, S., Kutyniok, G., Michaeli, T., Nerdel, C., Pfeffer, J., Poquet, O., Sailer, M., Schmidt, A., Seidel, T., & Kasneci, G. (2023). ChatGPT for good? On opportunities and challenges of large language models for education. *Learning and Individual Differences*, *103*, 102274. https://doi.org/10.1016/j.lindif.2023.102274

Koc, F. seyma, & Savas, P. (2025). The use of artificially intelligent chatbots in English language learning: A systematic meta-synthesis study of articles published between 2010 and 2024. *ReCALL*, *37*(1), 4–21. https://doi.org/10.1017/S0958344024000168

Kohen-Vacs, D., Usher, M., & Jansen, M. (2025). Integrating Generative AI into Programming Education: Student Perceptions and the Challenge of Correcting AI Errors. *International Journal of Artificial Intelligence in Education*. https://doi.org/10.1007/s40593-025-00496-4

Krathwohl, D. R. (2002). A Revision of Bloom's Taxonomy: An Overview. *Theory Into Practice*, *41*(4), 212–218. https://doi.org/10.1207/s15430421tip4104_2

Kuhail, M. A., Alturki, N., Alramlawi, S., & Alhejori, K. (2023). Interacting with educational chatbots: A systematic review. *Education and Information Technologies*, *28*(1), 973–1018. https://doi.org/10.1007/s10639-022-11177-3

Labadze, L., Grigolia, M., & Machaidze, L. (2023). Role of AI chatbots in education: Systematic literature review. *International Journal of Educational Technology in Higher Education*, *20*(1), 56. https://doi.org/10.1186/s41239-023-00426-1

Lally, A., & Fordor, P. (2011). Natural Language Processing With Prolog in the IBM Watson System. *The Association for Logic Programming (ALP) Newsletter*, *9*.

Lee, D., Arnold, M., Srivastava, A., Plastow, K., Strelan, P., Ploeckl, F., Lekkas, D., & Palmer, E. (2024). The impact of generative AI on higher education learning and teaching: A study of educators' perspectives. *Computers and Education: Artificial Intelligence*, *6*, 100221. https://doi.org/10.1016/j.caeai.2024.100221

Lin, Y., & Yu, Z. (2023). A bibliometric analysis of artificial intelligence chatbots in educational contexts. *Interactive Technology and Smart Education*, *ahead-of-print*(ahead-of-print). https://doi.org/10.1108/ITSE-12-2022-0165

Liu, G. L., Darvin, R., & Ma, C. (2024). Unpacking the role of motivation and enjoyment in AI-mediated informal digital learning of English (AI-IDLE): A mixed-method investigation in the Chinese context. *Computers in Human Behavior*, *160*, 108362. https://doi.org/10.1016/j.chb.2024.108362

Lo, C. K., Hew, K. F., & Jong, M. S. (2024). The influence of ChatGPT on student engagement: A systematic review and future research agenda. *Computers & Education*, *219*, 105100. https://doi.org/10.1016/j.compedu.2024.105100

Lytras, M., Visvizi, A., Damiani, E., & Mathkour, H. (2019). The cognitive computing turn in education: Prospects and application. In *Computers in Human Behavior* (pp. 446–449).

Ma, W., Ma, W., Hu, Y., & Bi, X. (2024). The who, why, and how of ai-based chatbots for learning and teaching in higher education: A systematic review. *Education and Information Technologies*. https://doi.org/10.1007/s10639-024-13128-6

Maheshwari, G. (2023). Factors influencing students' intention to adopt and use ChatGPT in higher education: A study in the Vietnamese context. *Education and Information Technologies*. https://doi.org/10.1007/s10639-023-12333-z

Mangal, M., & Pardos, Z. A. (2024). Implementing equitable and intersectionality-aware ML in education: A practical guide. *British Journal of Educational Technology*, *55*(5), 2003–2038. https://doi.org/10.1111/bjet.13484

Mittal, U., Sai, S., Chamola, V., et al. (2024). A comprehensive review on generative ai for education. *IEEE Access*.

Neyem, A., González, L. A., Mendoza, M., Alcocer, J. P. S., Centellas, L., & Paredes, C. (2024). Toward an AI Knowledge Assistant for Context-Aware Learning Experiences in Software Capstone Project Development. *IEEE Transactions on Learning Technologies*, *17*, 1599–1614. https://doi.org/10.1109/TLT.2024.3396735

Ng, D. T. K., Leung, J. K. L., Chu, S. K. W., & Qiao, M. S. (2021). Conceptualizing AI literacy: An exploratory review. *Computers and Education: Artificial Intelligence*, *2*, 100041. https://doi.org/10.1016/j.caeai.2021.100041

Ng, M. Y., Helzer, J., Pfeffer, M. A., Seto, T., & Hernandez-Boussard, T. (2025). Development of secure infrastructure for advancing generative artificial intelligence research in healthcare at an academic medical center. *Journal of the American Medical Informatics Association*, *32*(3), 586–588. https://doi.org/10.1093/jamia/ocaf005

OpenAI. (2022). *Introducing ChatGPT*. https://openai.com/index/chatgpt/

Pachava, V., Lasekan, O. A., Méndez-Alarcón, C. M., Pena, M. T. G., & Golla, S. K. (2025). Advancing SDG 4: Harnessing Generative AI to Transform Learning, Teaching, and Educational Equity in Higher Education. *Journal of Lifestyle and SDGs Review*, *5*(2), e03774–e03774. https://doi.org/10.47172/2965-730X.SDGsReview.v5.n02.pe03774

Pan, M., Lai, C., & Guo, K. (2025). Effects of GenAI-empowered interactive support on university EFL students' self-regulated strategy use and engagement in reading. *The Internet and Higher Education*, *65*, 100991. https://doi.org/10.1016/j.iheduc.2024.100991

Rahman, L. (2022). Cognitive computing and education and learning. *Int. J. Trend Sci. Res. Dev.(IJTSRD)*, *6*(3), 1628–1632.

Smutny, P., & Schreiberova, P. (2020). Chatbots for learning: A review of educational chatbots for the Facebook Messenger. *Computers & Education*, *151*, 103862. https://doi.org/10.1016/j.compedu.2020.103862

Stadler, M., Bannert, M., & Sailer, M. (2024). Cognitive ease at a cost: LLMs reduce mental effort but compromise depth in student scientific inquiry. *Computers in Human Behavior*, *160*, 108386. https://doi.org/10.1016/j.chb.2024.108386

Tzirides, A., Zapata, G., Bolger, P., Cope, B., Kalantzis, M., & Searsmith, D. (2024). Exploring Instructors' Views on Fine-Tuned Generative AI Feedback in Higher Education. *International Jl. On E-Learning*, *23*(3), 319–334. https://www.learntechlib.org/primary/p/225173/

Ulla, M. B., Advincula, M. J. C., Mombay, C. D. S., Mercullo, H. M. A., Nacionales, J. P., & Entino-Señorita, A. D. (2024). How can GenAI foster an inclusive language classroom? A critical language pedagogy perspective from Philippine university teachers. *Computers and Education: Artificial Intelligence*, *7*, 100314. https://doi.org/10.1016/j.caeai.2024.100314

UNESCO. (2022). *Recommendation on the ethics of artifical intelligence*. https://unesdoc.unesco.org/ark:/48223/pf0000381137

Urban, M., Děchtěrenko, F., Lukavský, J., Hrabalová, V., Svacha, F., Brom, C., & Urban, K. (2024). ChatGPT improves creative problem-solving performance in university students: An experimental study. *Computers & Education*, *215*, 105031. https://doi.org/10.1016/j.compedu.2024.105031

Wang, Y., & Xue, L. (2024). Using AI-driven chatbots to foster Chinese EFL students' academic engagement: An intervention study. *Computers in Human Behavior*, *159*, 108353. https://doi.org/10.1016/j.chb.2024.108353

Watermeyer, R., Phipps, L., Lanclos, D., & Knight, C. (2023). Generative AI and the automating of academia. *Postdigital Science and Education*. https://doi.org/10.1007/s42438-023-00440-6

Weizenbaum, J. (1983). ELIZA — a computer program for the study of natural language communication between man and machine. *Commun. ACM*, *26*(1), 23–28. https://doi.org/10.1145/357980.357991

Wu, R., & Yu, Z. (2024). Do AI chatbots improve students learning outcomes? Evidence from a meta-analysis. *British Journal of Educational Technology*, *55*(1). https://doi.org/10.1111/bjet.13334

Yang, Z., Wu, J. G., & Xie, H. (2024). Taming Frankenstein's monster: Ethical considerations relating to generative artificial intelligence in education. *Asia Pacific Journal of Education*, 1–14. https://doi.org/10.1080/02188791.2023.2300137

Yusuf, A., Pervin, N., & Román-González, M. (2024). Generative AI and the future of higher education: A threat to academic integrity or reformation? Evidence from multicultural perspectives. *International Journal of Educational Technology in Higher Education*, *21*(1), 21. https://doi.org/10.1186/s41239-024-00453-6

Zaim, M., Arsyad, S., Waluyo, B., Ardi, H., Al Hafizh, Muhd., Zakiyah, M., Syafitri, W., Nusi, A., & Hardiah, M. (2024). AI-powered EFL pedagogy: Integrating generative AI into university teaching preparation through UTAUT and activity theory. *Computers and Education: Artificial Intelligence*, *7*, 100335. https://doi.org/10.1016/j.caeai.2024.100335

Zhang, R., Zou, D., Cheng, G., & Xie, H. (2025). Flow in ChatGPT-based logic learning and its influences on logic and self-efficacy in English argumentative writing. *Computers in Human Behavior*, *162*, 108457. https://doi.org/10.1016/j.chb.2024.108457

Zhuhadar, L., & Ciampa, M. (2019). Leveraging learning innovations in cognitive computing with massive data sets: Using the offshore Panama papers leak to discover patterns. *Computers in Human Behavior*, *92*, 507–518. https://doi.org/10.1016/j.chb.2017.12.013

# Statements & Declarations

## Author Information

(Corresponding Author) Da-Lun Chen
Faculty of Information Technology and Electrical Engineering, University of Oulu, Oulu, Finland
da-lun.chen@oulu.fi

Prasasthy Balasubramanian
Center for Ubiquitous Computing, Faculty of Information Technology and Electrical Engineering, University of Oulu, Oulu, Finland
Prasasthy.Balasubramanian@oulu.fi

Lauri Lovén
Center for Ubiquitous Computing, Faculty of Information Technology and Electrical Engineering, University of Oulu, Oulu, Finland
Lauri.Loven@oulu.fi

Susanna Pirttikangas
Center for Ubiquitous Computing, Faculty of Information Technology and Electrical Engineering, University of Oulu, Oulu, Finland
susanna.pirttikangas@oulu.fi

Jaakko Sauvola
Faculty of Information Technology and Electrical Engineering, University of Oulu, Oulu, Finland
Jaakko.Sauvola@oulu.fi

Panagiotis Kostakos
Faculty of Information Technology and Electrical Engineering, University of Oulu, Oulu, Finland
Panos.Kostakos@oulu.fi

## Funding

The authors declare that no funds, grants, or other support were received during the preparation of this manuscript.

## Competing Interests

The authors declare that there are no known conflicting financial interests or personal relationships that may seem to have impacted on the findings presented in this paper.

## Author Contributions

DC: Research conception, literature review, survey design, data collection, data analysis, interpretation, and writing of all sections except the framework sections and 6.3.

PB: Research conception, survey design, data collection, writing of the framework sections, 6.3, and contributions to the initial literature review and related work.

LL, SP, JS: Research conception and survey design review.

PK: Research conception, survey design review, and manuscript editing.

All authors read and approved the final manuscript.

## Data availability

The data supporting this study's findings are available from the corresponding author upon reasonable request.